%% file: main.tex
\documentclass[11pt]{article}
\usepackage{xr}
\usepackage{hyperref}

\usepackage{graphicx}
\usepackage{amsmath}
\usepackage{amssymb} 
\usepackage[numbers]{natbib}

\usepackage{cleveref}
\usepackage{booktabs} 
\usepackage{multirow} 
\usepackage{subcaption}

\usepackage{bm}

\usepackage[margin=1in]{geometry}
\newcommand{\bftab}{\fontseries{b}\selectfont}

\begin{document}

\title{Optimising antibiotic switching via forecasting of patient physiology}
\author{ 
    Magnus Ross$^1$\thanks{Email: \texttt{magnus.ross@ucl.ac.uk, \texttt{v.lampos@ucl.ac.uk}}}, 
    Nel Swanepoel$^2$, 
    Akish Luintel$^3$, 
    Emma McGuire$^3$, \\
    Ingemar J. Cox$^{1,4}$, 
    Steve Harris$^2$, 
    Vasileios Lampos$^{1*}$ \\[1em]
    \small $^1$Department of Computer Science, Centre for AI, UCL, UK \\
    \small $^2$Institute of Health Informatics, UCL, UK\\
    \small $^3$Department of Medical Microbiology, Infection Division, University College Hospital London, UK\\
    \small $^4$Department of Computer Science, University of Copenhagen, Denmark
}
\date{} 

\maketitle

\begin{abstract}
Timely transition from intravenous (IV) to oral antibiotic therapy shortens hospital stays, reduces catheter-related infections, and lowers healthcare costs, yet one in five patients in England remain on IV antibiotics despite meeting switching criteria. Clinical decision support systems can improve switching rates, but approaches that learn from historical decisions reproduce the delays and inconsistencies of routine practice. We propose using neural processes to model vital sign trajectories probabilistically, predicting switch-readiness by comparing forecasts against clinical guidelines rather than learning from past actions, and ranking patients to prioritise clinical review. The design yields interpretable outputs, adapts to updated guidelines without retraining, and preserves clinical judgement. Validated on MIMIC-IV (US intensive care, 6{,}333 encounters) and UCLH (a large urban academic UK hospital group, 10{,}584 encounters), the system selects 2.2--3.2$\times$ more relevant patients than random. Our results demonstrate that forecasting patient physiology offers a principled foundation for decision support in antibiotic stewardship.
\end{abstract}

\section{Introduction}

The timely transition of hospital patients from intravenous (IV) to oral antimicrobial therapy, known as the IV-to-oral switch (IVOS), benefits patients and hospitals~\citep{cyriac_switch_2014}. Clinically, it can shorten hospital stays and reduce catheter-related bloodstream infections~\citep{wald-dickler_oral_2022, zanella_dwell_2025}; operationally, it can free nursing time and reduce costs~\citep{jenkins_iv_2023}. Yet IV therapy often continues after an oral alternative would be safe. In England, 19\% of patients remain on IV antibiotics despite meeting switching criteria~\citep{Fingertips2025}, and some are not switched at all~\citep{li_oral_2019}. IVOS depends on timely review of a changing clinical picture, but that review is easy to miss on busy wards, especially when senior input or microbiology advice is delayed.
\par
IVOS decisions have two linked components~\citep{ukhsa_guidelines,harvey_criteria_2023}. The first is \emph{clinical stability}: whether the patient is improving and likely to remain stable, assessed using trends in vital signs (e.g., temperature, heart rate, blood pressure, respiratory rate, oxygen saturation) and, where available, laboratory markers. The second is \emph{oral suitability}: whether oral therapy is appropriate and feasible, for example that the patient can tolerate oral intake and that an appropriate oral agent is available. This paper targets clinical stability. Our task is to identify patients whose recent observations imply a high probability of staying within normal ranges in the near future, so they can be prioritised for review.
\par
This problem has a practical constraint~\cite{10.1093/ofid/ofaf721}. In England, the median hospital has around 550 general/acute/maternity inpatients at any time, and about one third receive antibiotics. That volume exceeds the capacity of typical antimicrobial stewardship teams (infection specialists and specialist pharmacists). Decision support must therefore help them focus on the patients most likely to benefit from review. IVOS decisions also depend on information that is hard to capture reliably in structured EHR data (for example, surgical wound progress, source control plans, evolving imaging, or bedside assessments). The goal of a CDSS is therefore not to automate IVOS, but to surface patients for expert review and discussion with the primary team.
\par
Clinical decision support systems (CDSSs) can increase switching rates and reduce IV therapy days. Prior rule-based systems have reduced IV use through alerts based on fixed criteria~\citep{beeler_earlier_2015, berrevoets_electronic_2017, kan_implementation_2019, quintens_basic_2022} (see Supplementary \labelcref{sec:supp_rule_based} for a detailed review). However, fixed rules struggle to assess whether a patient is \emph{improving} rather than merely meeting thresholds at one point in time. AI-enabled methods that learn from historical switching behaviour face a deeper limitation: if the training signal is what clinicians \emph{did}, then the best the model can do is reproduce that sub-optimal behaviour. It cannot reliably improve practice against a clinical standard~\citep{bolton_personalising_2024,ross2025expediting}.
\par
We propose an alternative approach that forecasts patient physiology rather than past clinical actions. Using neural processes to model vital sign trajectories probabilistically, we predict switch-readiness by applying clinical criteria to these forecasts. This avoids learning historical delays, side-steps the need to label switches from electronic health records (EHRs), and yields outputs clinicians can inspect directly (the forecast trajectories and their uncertainty). Because criteria are applied after forecasting, the system can be updated as clinical standards evolve without retraining the forecasting model, and it can be adapted in non-standard cases (e.g., discounting a variable that is clinically uninformative in context such as fever in patients on antipyretics). The system ranks patients by predicted switch-readiness, helping teams prioritise reviews while preserving clinical judgement for factors outside routine vital signs data.
\par
\begin{figure}
    \centering
    
    \includegraphics[width=1\textwidth]{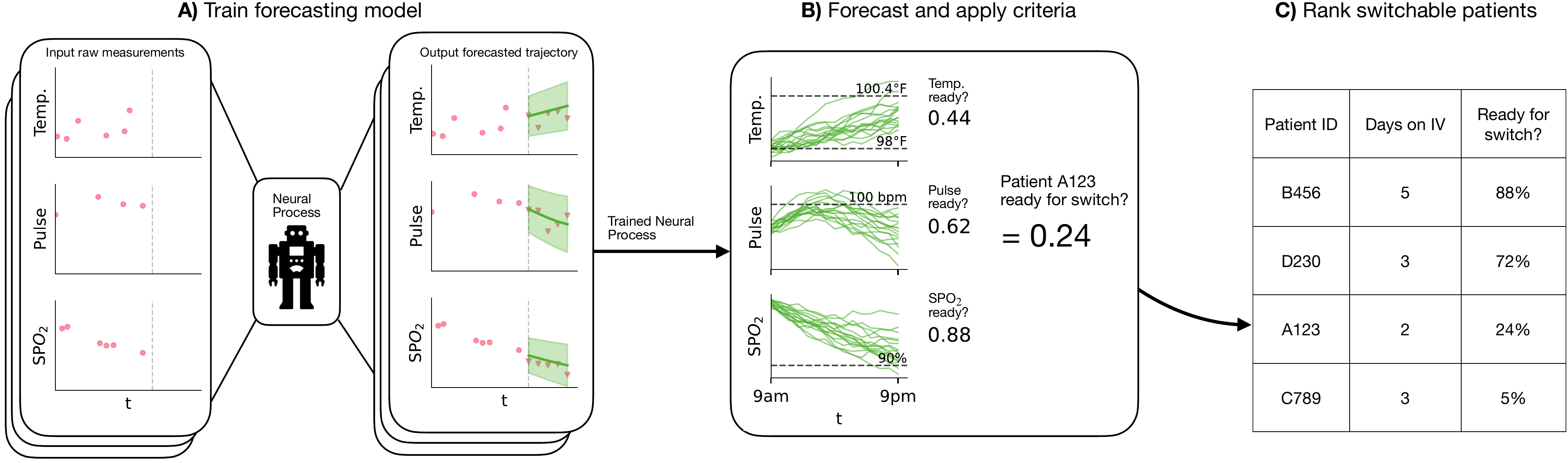}
    \caption{The proposed AI-enabled IVOS system. \textbf{A)} Training a probabilistic neural process model to forecast patient trajectories. \textbf{B)} Using the trained model to generate likely trajectories over the following day and applying clinical criteria to determine the proportion of samples ready to be switched. \textbf{C)} Presentation of switch-readiness probabilities as a ranked list of patients for review.}
    \label{fig:main_diagram}
\end{figure}
\Cref{fig:main_diagram} illustrates our system architecture. We forecast key vital signs (\textbf{A}), apply clinical criteria to estimate switch-readiness probabilities (\textbf{B}), and use these to rank patients for review (\textbf{C}). Oral suitability and the final IVOS decision remain with clinicians, who consider the broader clinical record and coordinate with the primary team after reviewing flagged patients~\citep{ukhsa_guidelines,harvey_criteria_2023}.

\begin{table}
    \centering

    \renewcommand{\arraystretch}{1.3}
    \resizebox{\columnwidth}{!}{
    \begin{tabular}{p{6cm}ll} 
        \toprule
     \textbf{Statistic} & \textbf{MIMIC} & \textbf{UCLH} \\
        \midrule
        Number of admissions & 6,333 & 10,584 \\
        Unique subjects & 5,850 & 8,701 \\
        Age (years) & 65.70 $\pm$ 15.29 & 55.98 $\pm$ 19.24 \\
        Length of stay (days) & 17.69 $\pm$ 17.07 & 14.23 $\pm$ 14.42 \\
        \hline
        \textbf{Sex} & & \\
        \hspace{3mm} Female & 2,579 (40.7\%) & 5,564 (52.6\%) \\
        \hspace{3mm} Male & 3,754 (59.3\%) & 5,019 (47.4\%) \\
        \hline
        \textbf{Top 5 Race/Ethnicity} & & \\
        & White: 3,772 (59.6\%) & White British: 3,324 (31.4\%) \\
        & Unknown: 1,074 (17.0\%) & Unknown: 3,033 (28.7\%) \\
        & Black: 720 (11.4\%) & Other White Bg.: 1,204 (11.4\%) \\
        & Other: 292 (4.6\%) & Other Ethnic Group: 671 (6.3\%) \\
        & Hispanic: 247 (3.9\%) & Black African: 509 (4.8\%) \\
        \hline
        \textbf{Top 5 Oral Antibiotics} & & \\
        & Azithromycin: 724 (11.4\%) & Co-amoxiclav: 3,471 (32.8\%) \\
        & Metronidazole: 704 (11.1\%) & Clarithromycin: 797 (7.5\%) \\
        &  Sulfameth/Trimeth: 477 (7.5\%) & Ciprofloxacin: 771 (7.3\%) \\
        & Doxycycline: 371 (5.9\%) & Amoxicillin: 421 (4.0\%) \\
        & Vancomycin: 370 (5.8\%) & Doxycycline: 410 (3.9\%) \\
        \hline
        \textbf{Top 5 IV Antibiotics} & & \\
        & Vancomycin: 4,158 (65.7\%) & Co-amoxiclav: 4,920 (46.5\%) \\
        & Cefepime: 2,774 (43.8\%) & Pip.-Taz.: 3,841 (36.3\%) \\
        & Ceftriaxone: 2,629 (41.5\%) & Metronidazole: 2,849 (26.9\%) \\
        & Cefazolin: 2,249 (35.5\%) & Meropenem: 1,208 (11.4\%) \\
        & Metronidazole: 1,449 (22.9\%) & Ciprofloxacin: 882 (8.3\%) \\
        \hline
        \textbf{Measurement variable} & \textit{Mean \# per admission} & \textit{Mean \# per admission} \\
        \hspace{3mm} Heart rate & 157.8 & 81.1 \\
        \hspace{3mm} Respiratory rate & 154.0 & 78.4 \\
        \hspace{3mm} SpO$_2$ & 154.9 & 80.8 \\
        \hspace{3mm} Systolic BP & 84.0 & 62.9 \\
        \hspace{3mm} Temperature & 36.1 & 61.9 \\
        \bottomrule
    \end{tabular}
    }
        \caption{Comparison of the evaluation set cohorts for MIMIC-IV and UCLH datasets. Categorical variables are reported as count (percentage of admissions). Continuous variables are reported as mean $\pm$ standard deviation. MIMIC-IV records race while UCLH records ethnicity, reflecting the coding systems used in each source database; ``Other White Bg'' denotes Other White Background. Antibiotic frequencies may sum to more than 100\% as patients can receive multiple antibiotics during an admission. The bottom section shows the mean number of vital sign measurements recorded per admission for each variable. Abbreviations: Pip.-Taz. (Piperacillin-Tazobactam), Sulfameth/Trimeth (Sulfamethoxazole-Trimethoprim), IV (Intravenous), SpO$_2$ (Oxygen Saturation), BP (Blood Pressure).}
    \label{tab:cohort_characteristics}
\end{table}

\section{Results}

We validate our method using two distinct EHR datasets: MIMIC-IV v3.1 \citep{johnson_mimic-iv_2023}, comprising ICU admissions from Beth Israel Deaconess Medical Center in Boston, USA, and a non-public dataset from University College London Hospitals (UCLH), a general hospital group in London, UK. These cohorts represent significantly different populations and clinical practices, providing a robust test of model generalisability (see \Cref{tab:cohort_characteristics} for detailed cohort characteristics).

The core task of our CDSS is to support daily IVOS reviews by predicting patient physiological stability. Unless otherwise stated, we forecast five key vital signs---heart rate, respiratory rate, oxygen saturation, systolic blood pressure, and temperature---over a 12-hour future window. A patient is defined as ``switch-ready'' if all forecast vital signs remain within clinically normal ranges throughout this window. We model these trajectories using Neural Processes (NPs) \citep{garnelo_conditional_2018}, a class of probabilistic models chosen for their ability to handle the irregular sampling and uncertainty inherent in clinical time-series data. Detailed model architecture and training procedures are provided in \Cref{sec:methods_np}.

\subsection{Forecasting}
\label{sec:results_forecasting}
\begin{table}
\centering
\resizebox{\textwidth}{!}{
\begin{tabular}{llccccc}
\toprule

\multirow{3}{*}{\textbf{Dataset}} & \multirow{3}{*}{\textbf{Model}} & \multicolumn{5}{c}{\textbf{Variable}} \\
\cmidrule(lr){3-7}
& & \textbf{HR} & \textbf{RR} & \textbf{SpO}$_2$ & \textbf{SBP} & \textbf{Temp.} \\
& & (bpm) & (bpm) & (\%) & (mmHg) & ($^{\circ}$F) \\
\midrule

\multirow{3}{*}{\textbf{UCLH}}
& \textbf{Repeat} & 8.79 [8.76-8.82] & 1.97 [1.96-1.98] & 1.51 [1.51-1.52] & 12.86 [12.81-12.91] & 0.69 [0.69-0.70] \\
& \textbf{GBDT}   & 7.94 [7.92-7.96] & 1.76 [1.76-1.77] & \bftab 1.31 [1.30-1.31] & 10.86 [10.83-10.89] & 0.58 [0.58-0.58] \\
& \textbf{NP}     & \bftab 7.58 [7.56-7.60] & \bftab 1.71 [1.70-1.72] & \bftab 1.30 [1.30-1.31] & \bftab 10.71 [10.67-10.75] & \bftab 0.57 [0.57-0.57] \\
\midrule

\multirow{3}{*}{\textbf{MIMIC}}
& \textbf{Repeat} & 9.17 [9.14-9.19] & 4.09 [4.08-4.11] & 2.05 [2.04-2.05] & 14.29 [14.24-14.34] & 0.63 [0.62-0.63] \\
& \textbf{GBDT}   & 8.92 [8.90-8.94] & 3.49 [3.48-3.50] & 1.79 [1.78-1.79] & 12.27 [12.24-12.30] & 0.55 [0.55-0.55] \\
& \textbf{NP}     & \bftab 8.32 [8.29-8.34] & \bftab 3.40 [3.39-3.41] & \bftab 1.76 [1.76-1.77] & \bftab 11.97 [11.92-12.01] & \bftab 0.54 [0.54-0.54] \\
\bottomrule
\end{tabular}
}
\caption{
Mean absolute error of model forecasts for 12 hour prediction horizon. The 95\% confidence intervals are shown, computed by a bootstrap resampling of the mean error over each of the windows in the test set. Bold values indicate the best performing model(s) (p $\ge$ 0.05).
Variable abbreviations: HR (Heart Rate), RR (Respiratory Rate), $\text{SpO}_2$ (Oxygen Saturation), SBP (Systolic Blood Pressure), Temp. (Temperature). Model abbreviations: GBDT (Gradient Boosted Decision Trees), NP (Neural Process).
}
\label{tab:forecasting-mae}
\end{table}
We first aim to establish the forecasting performance of the proposed neural process model. \Cref{tab:forecasting-mae} shows the point forecasting performance of the proposed NP model, and two baselines. The model is tasked to predict the values of any measurements occurring in the 12-hour period immediately following the prediction time, defined as the point up to which patient history is known. The repeat baseline simply uses the last measured value prior to the forecast window as the prediction. The gradient boosted decision tree is more sophisticated, and uses a set of statistical features generated from past measurements \citep{ke_lightgbm_2017}. For discussion of baseline construction, see \Cref{sec:methods_baslines}. The NP model provides a significant improvement over both baseline models on all measurements except $\text{SpO}_2$, although the GBDT scores are only marginally worse. The mean improvement in the NP scores relative to the GBDT across variables was 2.25\% for UCLH and 2.99\% for MIMIC. The UCLH data is more easily forecastable than the MIMIC data, reflected in the lower errors for all models in each measurement. This is likely due to the fact that the MIMIC ICU cohort is made up of patients with more serious, less stable conditions. 
\par
\begin{figure}
    \centering
    \includegraphics[width=1.0\textwidth]{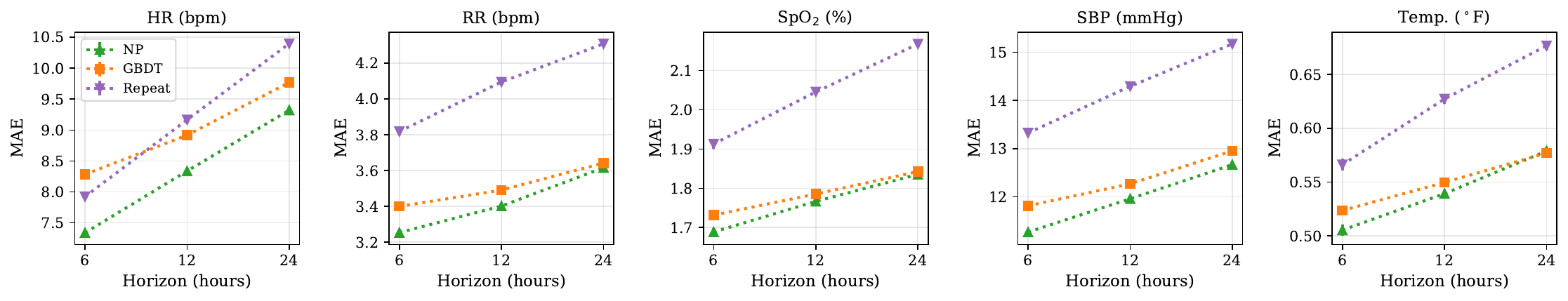}
    \caption{Forecasting performance on the MIMIC data for each variable with increasing forecast horizon for NP model and baselines. Note horizontal axis scale is logarithmic. Error bars show 95\% confidence. Variable Abbreviations: HR (Heart Rate), RR (Respiratory Rate), $\text{SpO}_2$ (Oxygen Saturation), SBP (Systolic Blood Pressure), Temp. (Temperature).  Model abbreviations: GBDT (Gradient Boosted Decision Trees), NP (Neural Process).}
    \label{fig:mae_horizon_comparison}
\end{figure}

Although a 12 hour forecast horizon is most appropriate for our clinical context, this is likely to vary across institutions with different workflows. As such, we also validate the model performance on two additional horizons: 6 hours and 1 day. \Cref{fig:mae_horizon_comparison} shows the result of this experiment for the MIMIC data. As expected, the mean absolute error increases as the forecast horizon increases; points further into the future are forecasted with lower accuracy. For the shorter horizons, the NP is significantly better than all baselines across variables. As the horizon increases the performance of the GBDT and NP converge, with the models providing similar performance on the 1 day horizon. We note that the NP model is probabilistic, and as such produces a forecasting distribution over possible measurement values as opposed to a single point estimate like the baseline models. This probabilistic output enables its predictions to be used downstream for the purposes of ranking, patient's readiness for switching, as we will see in the following section. \Cref{tab:forecasting-mae} does not assess this probabilistic aspect of the model and quantifies only the point forecasting performance. 

\subsection{Switch-readiness prediction}
\label{sec:results_switch_readiness}
The key machine learning task in the proposed CDSS is the prediction of a patient's readiness for switching in the forecast window. We use a set of clinical criteria, which define a normal range for each of the key vital measurements, as a proxy for switch readiness. If a patient's measurements fall within the normal range over the forecast window, we label them as ready to be switched on that day. The two datasets have disparate class distributions, reflecting the differences in the populations: in the UCLH data patients are ready to be switched on 40.2\% of the test days, compared to only 12.2\% in the more unwell MIMIC ICU population.
\par
The NP model produces a distribution over predicted measurements. We use this distribution to determine the probability a patient will meet the criteria in the forecast window, and thus the probability they are ready to be switched. Patients with a higher probability of switch-readiness are ranked higher in the list presented to clinicians, and thus are prioritised for review. The two-step process of first forecasting vitals, then applying criteria, gives an interpretable output that clinicians can use to inform their decision making, since the prediction of the patient's state underlying the predicted probability is clearly understandable. 
\par
An alternative, less interpretable, approach is to directly predict the switch readiness label using a supervised classification model.  We apply the direct prediction method to serve as baseline. As inputs to the baseline models we use a set of statistical features generated from the past measurements. We apply both logistic regression and gradient boosted decision trees. For the NP model, we use two variants, the standard NP, as described above, and a variant tuned to produce well calibrated probabilities (NP-Tuned). This variant uses an additional classification head that operates on the encodings generated by the final layer of the forecasting model to directly predict switch readiness. The issue of calibration is discussed further in Supplementary \Cref{sec:supp_calibration}.
\par
\begin{table}
\centering
\resizebox{\columnwidth}{!}{
\begin{tabular}{llrrrr}
\toprule
\textbf{Dataset} & \textbf{Model} & \multicolumn{4}{c}{\textbf{Metric}} \\
\cmidrule(lr){3-6}
& & \textbf{AUROC} & \textbf{AP} & \textbf{Brier Score} & \textbf{Precision@5} \\
\midrule
\multirow{4}{*}{\textbf{UCLH}} & \textbf{Logistic} & 0.827 [0.823-0.831] & 0.740 [0.732-0.747] & 0.167 [0.165-0.169] & 0.836 [0.823-0.850] \\
& \textbf{GBDT} & 0.862 [0.858-0.865] & \bftab 0.794 [0.787-0.800] & 0.149 [0.147-0.151] &\bftab 0.899 [0.888-0.910] \\
& \textbf{NP} & 0.858 [0.854-0.861] & 0.785 [0.779-0.792] & 0.216 [0.212-0.219] & 0.883 [0.871-0.894] \\
& \textbf{NP-tuned} & \bftab 0.867 [0.864-0.871] & \bftab 0.800 [0.793-0.807] & \bftab 0.146 [0.144-0.148] & \bftab 0.901 [0.890-0.912] \\
\midrule
\multirow{4}{*}{\textbf{MIMIC}} & \textbf{Logistic} & 0.759 [0.749-0.767] & 0.300 [0.284-0.315] & 0.098 [0.095-0.100] & 0.280 [0.263-0.297] \\
& \textbf{GBDT} & 0.869 [0.863-0.875] & \bftab 0.471 [0.453-0.490] & \bftab 0.081 [0.079-0.084] &\bftab 0.381 [0.363-0.400] \\
& \textbf{NP} & \bftab 0.882 [0.877-0.887] & \bftab 0.477 [0.459-0.497] & 0.106 [0.102-0.110] & \bftab 0.393 [0.373-0.413] \\
& \textbf{NP-tuned} & 0.862 [0.856-0.868] & 0.453 [0.433-0.472] & \bftab 0.083 [0.080-0.085] & 0.369 [0.350-0.387] \\
\bottomrule
\end{tabular}
} 
\caption{Comparison of model performance for switch-readiness prediction. Higher values are better for all metrics except the Brier score. Bold values indicate the best performing model(s) (p $\ge$ 0.05). Metric abbreviations: AUROC (Area under the Receiver Operating Characteristic), AP (Average Precision). Model abbreviations: GBDT (Gradient Boosted Decision Trees), NP (Neural Process).}
\label{tab:switch_ranking_performance}
\end{table}
In order to evaluate the ability of the models to predict the readiness of the patients for switching, we use four metrics: AUROC, average precision (AP), Brier score, and precision@5 (see Supplementary \labelcref{sec:eval_metrics} for detailed definitions). These metrics are chosen to align with the intended deployment of the models: producing a ranked list of patients most likely to be ready for review. 
The AUROC gives a global measure of the model's ability to discriminate between the positive and negative classes, representing the probability that a randomly chosen positive patient is ranked higher than a randomly chosen negative one. The average precision is also reported because, unlike AUROC, it is particularly sensitive to the model's performance on the positive (switch-ready) class of which there is a lower proportion. A high average precision score indicates that the model populates the top of the ranked list of predicted probabilities with a high concentration of true positives.  The Brier score is included to assess the accuracy and calibration of the probability scores themselves; a low Brier score indicates the model's predicted probabilities reliably represent the true chance of the event occurring \citep{brier_verification_1950}. The precision@5 metric corresponds to the proportion of the model's top 5 ranked patients on each day that are truly ready to be switched, averaged over all days in the test set. This metric is intended to reflect the deployment context of the model, in which a ranked list will be presented to a clinician on a daily cadence. 

\Cref{tab:switch_ranking_performance} shows the performance of each model on the test set from both the UCLH and MIMIC data; statistical significance was determined via bootstrap resampling (see Supplementary \labelcref{sec:supp_significance}). These results indicate that both the GBDT baseline and the NP models are able to accurately rank patients on both datasets. A discussion of the advantages of the NP approach is provided in \Cref{sec:discussion}.  For a random model, we would expect the AUROC to be 0.5, and the average precision and precision@5 to be equal to the proportion of positive examples. The precision@5, the metric most closely aligned with the intended clinical deployment, indicates that the NP model would select $2.2 \times$ more relevant patients than random selection on the UCLH data, and $3.2 \times$ on the MIMIC data. On the UCLH data the NP-tuned model provides the highest scores across all metrics, although for the average precision and precision@5 metrics the GBDTs performance is statistically indistinguishable. On the MIMIC data, there is more dispersion: the standard NP model has a significantly better AUROC, whereas for the other metrics (AP, precision@5, Brier score) the performance of the GBDT and NP is similar. The poor calibration of the standard NP model on both datasets is indicated by the elevated Brier score; this issue is discussed further in Supplementary \labelcref{sec:supp_calibration}. As in the case of forecasting, we find that the UCLH data is more predictable. The strong performance of all models in selecting relevant patients validates the overall approach of ranking patients for IVOS review by forecasting physiology and applying clinical criteria, rather than learning from historical switching decisions.

\begin{figure}
    \centering

    \begin{subfigure}{0.9\textwidth}
        \centering
        \includegraphics[width=\textwidth]{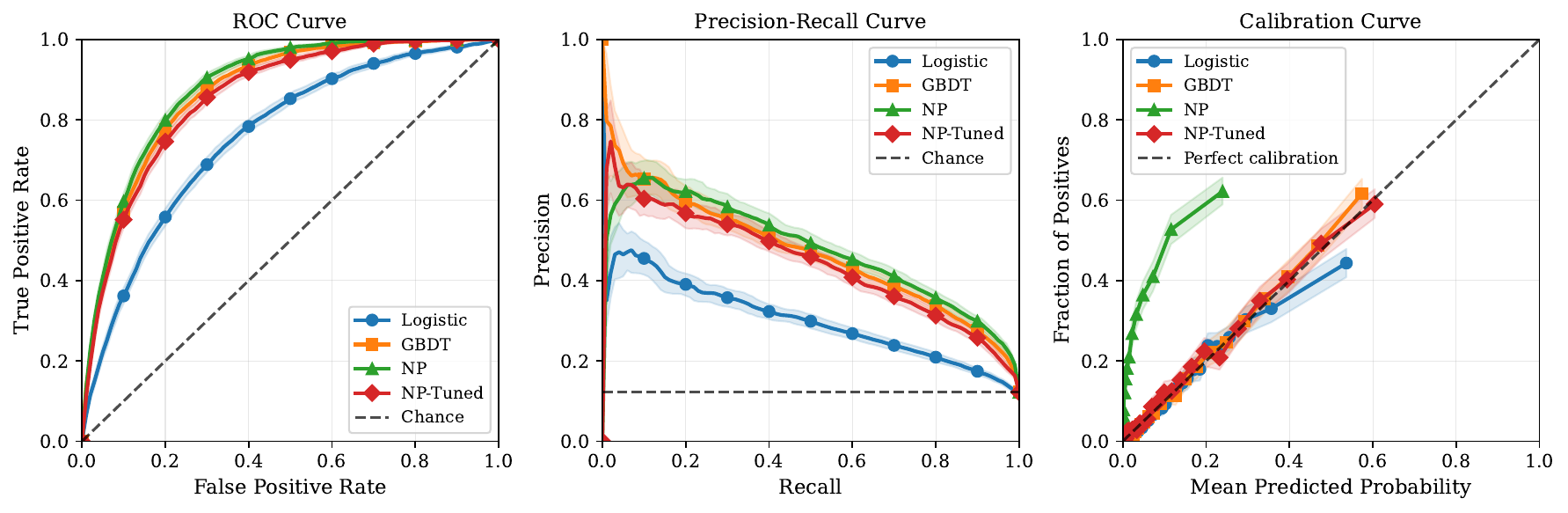}
        \caption{MIMIC}
        \label{fig:comparison_curves_mimic_sub}
    \end{subfigure}
    \hfill
    \begin{subfigure}{0.9\textwidth}
        \centering
        \includegraphics[width=\textwidth]{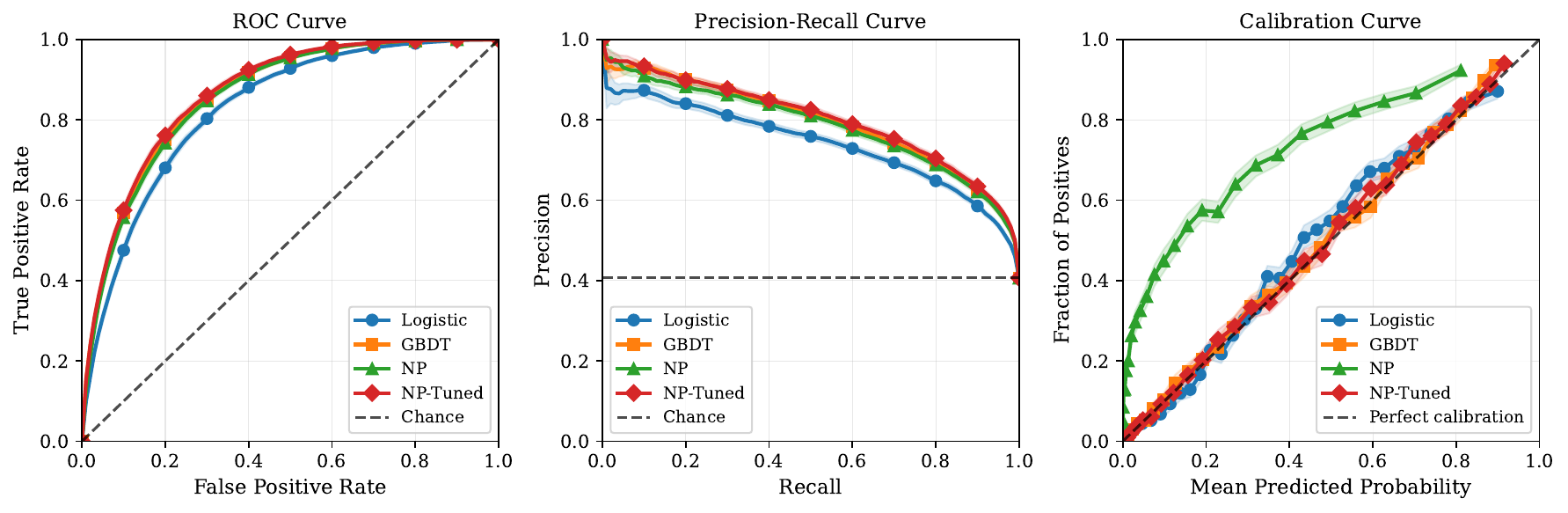}
        \caption{UCLH}
        \label{fig:comparison_curves_uclh_sub}
    \end{subfigure}
    
    \caption{Comparison of model performance curves for both datasets. From left to right the plots show Receiver Operating Characteristic, precision-recall, and calibration curves. For the former two the dashed line shows the performance of a chance model, for the latter it shows a perfect model. For the calibration curve, aggregation bins are the quantiles of the distribution of predicted probabilities; the small proportion of positive examples in MIMIC means the curves do not span the full range of the plot. The shaded region indicates 95\% confidence.  Model abbreviations: GBDT (Gradient Boosted Decision Trees), NP (Neural Process).}
    \label{fig:comparison_curves_combined}
\end{figure}

\begin{figure}
    \centering
    \includegraphics[width=1.0\textwidth]{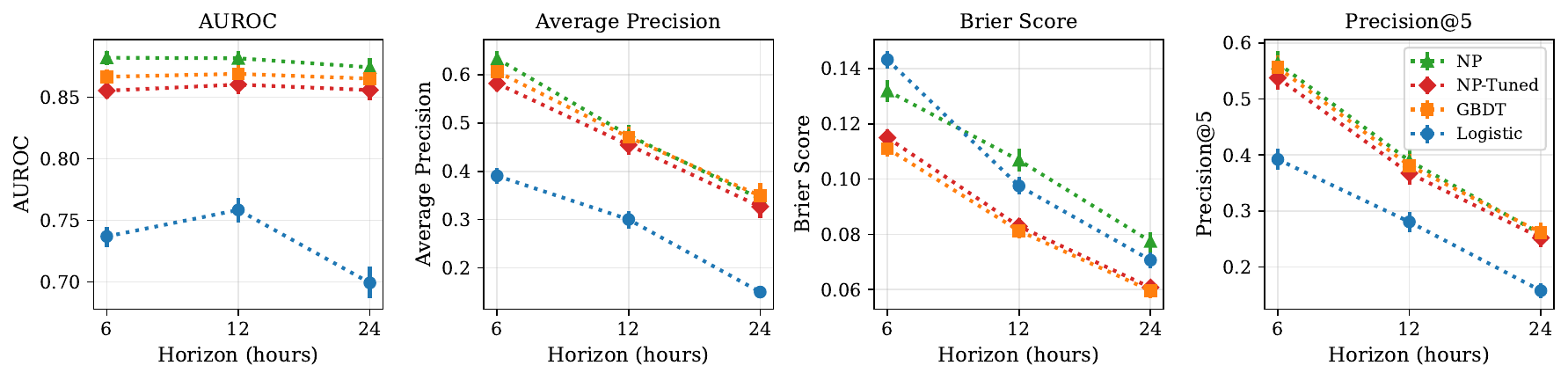}
    \caption{Ranking performance on the MIMIC data for each variable with increasing forecast horizon for NP models and baselines. Horizontal axis scale is logarithmic. Error bars show 95\% confidence. Note that the classes become more imbalanced as the forecast horizon increases, with 20.1\%, 12.2\%, and 7.8\% of the instances being positive for the 6, 12, and 24 hour horizons respectively.  Model abbreviations: GBDT (Gradient Boosted Decision Trees), NP (Neural Process).}
    \label{fig:ranking_horizon_comparison}
\end{figure}

\Cref{fig:comparison_curves_combined} shows performance curves for each of the models on both datasets. These curves again reinforce the finding that the GBDT and NP models perform similarly well in ranking patients on both the datasets, with the NP model providing a small improvement on MIMIC. The poor calibration of the standard NP is evident in the calibration curve, where the model consistently underestimates probabilities. All other models, including the tuned NP, are well calibrated on both datasets. We also validate the ranking performance of the NP models and baselines on varying forecast horizons, with results shown in \Cref{fig:ranking_horizon_comparison}. The ordering of the models remains consistent across forecasting horizons, with both the NP and GBDT providing strong performance. Note that the Brier score decreases as the forecast horizon increases, despite the problem being harder, because the fraction of positive instances decreases.

Our primary concern is the performance of models at the task of ranking patients as most ready for switching, as this is most closely aligned with the intended deployment of the models. Ranking does not require setting a binary decision threshold, and we do not intend to use the models to directly make a decision to switch or not. Nevertheless, it may still be instructive to compare the models' ability to predict this binary class. \Cref{tab:binary_prediction_performance} provides a number of metrics assessing the binary prediction performance of the models. A decision threshold was set for each model by computing the value that gave the highest F1 score on the validation set. On the UCLH data, both the GBDT and NP-tuned models achieve the best performance for accuracy and F1 score. On the MIMIC data, the standard NP achieves the best F1 score, whilst the Repeat and NP models both achieve the best recall. The GBDT baseline generally achieves significantly better precision. Consistent with the forecasting and ranking results, performance is lower on the MIMIC dataset, where the low prevalence of the positive class leads to significantly reduced precision across all models.

\begin{table}[ht]
\centering
\resizebox{\columnwidth}{!}{
\begin{tabular}{llrrrr}
\toprule
\textbf{Dataset} & \textbf{Model} & \multicolumn{4}{c}{\textbf{Metric}} \\
\cmidrule(lr){3-6}
& & \textbf{F1 Score} & \textbf{Accuracy} & \textbf{Precision} & \textbf{Recall} \\
\midrule
\multirow{5}{*}{\textbf{UCLH}} & \textbf{Repeat} & 0.704 [0.699-0.709] & 0.700 [0.696-0.704] & 0.587 [0.581-0.594] & \bftab 0.879 [0.874-0.884] \\
& \textbf{Logistic} & 0.719 [0.714-0.724] & 0.736 [0.732-0.740] & 0.634 [0.627-0.640] & 0.832 [0.827-0.838] \\
& \textbf{GBDT} &\bftab  0.744 [0.740-0.749] &\bftab 0.765 [0.761-0.769] & \bftab 0.667 [0.661-0.673] & 0.843 [0.837-0.848] \\
& \textbf{NP} & 0.741 [0.736-0.746] & 0.754 [0.751-0.759] & 0.649 [0.643-0.655] & 0.863 [0.857-0.868] \\
& \textbf{NP-tuned} & \bftab 0.748 [0.744-0.754] &\bftab 0.764 [0.760-0.768] & 0.660 [0.654-0.666] & 0.864 [0.859-0.870] \\
\midrule
\multirow{5}{*}{\textbf{MIMIC}} & \textbf{Repeat} & 0.432 [0.419-0.446] & 0.778 [0.773-0.784] & 0.315 [0.303-0.327] & \bftab 0.691 [0.673-0.708] \\
& \textbf{Logistic} & 0.372 [0.359-0.385] & 0.763 [0.758-0.769] & 0.275 [0.264-0.287] & 0.575 [0.557-0.594] \\
& \textbf{GBDT} & 0.497 [0.482-0.511] & \bftab 0.861 [0.856-0.865] &\bftab 0.444 [0.428-0.460] & 0.564 [0.545-0.581] \\
& \textbf{NP} & \bftab 0.519 [0.505-0.531] & 0.845 [0.840-0.850] & 0.418 [0.404-0.432] &\bftab 0.683 [0.666-0.701] \\
& \textbf{NP-tuned} & 0.494 [0.479-0.508] & 0.844 [0.839-0.848] & 0.409 [0.393-0.423] & 0.626 [0.608-0.644] \\
\bottomrule
\end{tabular}
} 
\caption{Comparison of model performance for binary classification, with the positive class representing a patient that is ready to be switched. The proportion of positive instances is 40.2\% and 12.2\% for UCLH and MIMIC respectively, indicating predicting the majority class could achieve an accuracy of 59.8\% and 87.8\% in each case. Decision thresholds were chosen to maximise the F1 score on the validation set. 95\% confidence intervals are shown in brackets. Bold values indicate the best performing model(s) (p $\ge$ 0.05). A higher value is better for all metrics. Model abbreviations: GBDT (Gradient Boosted Decision Trees), NP (Neural Process).}
\label{tab:binary_prediction_performance}
\end{table}

\section{Discussion}
\label{sec:discussion}

\begin{figure}
    \centering
    \includegraphics[width=1.0\textwidth]{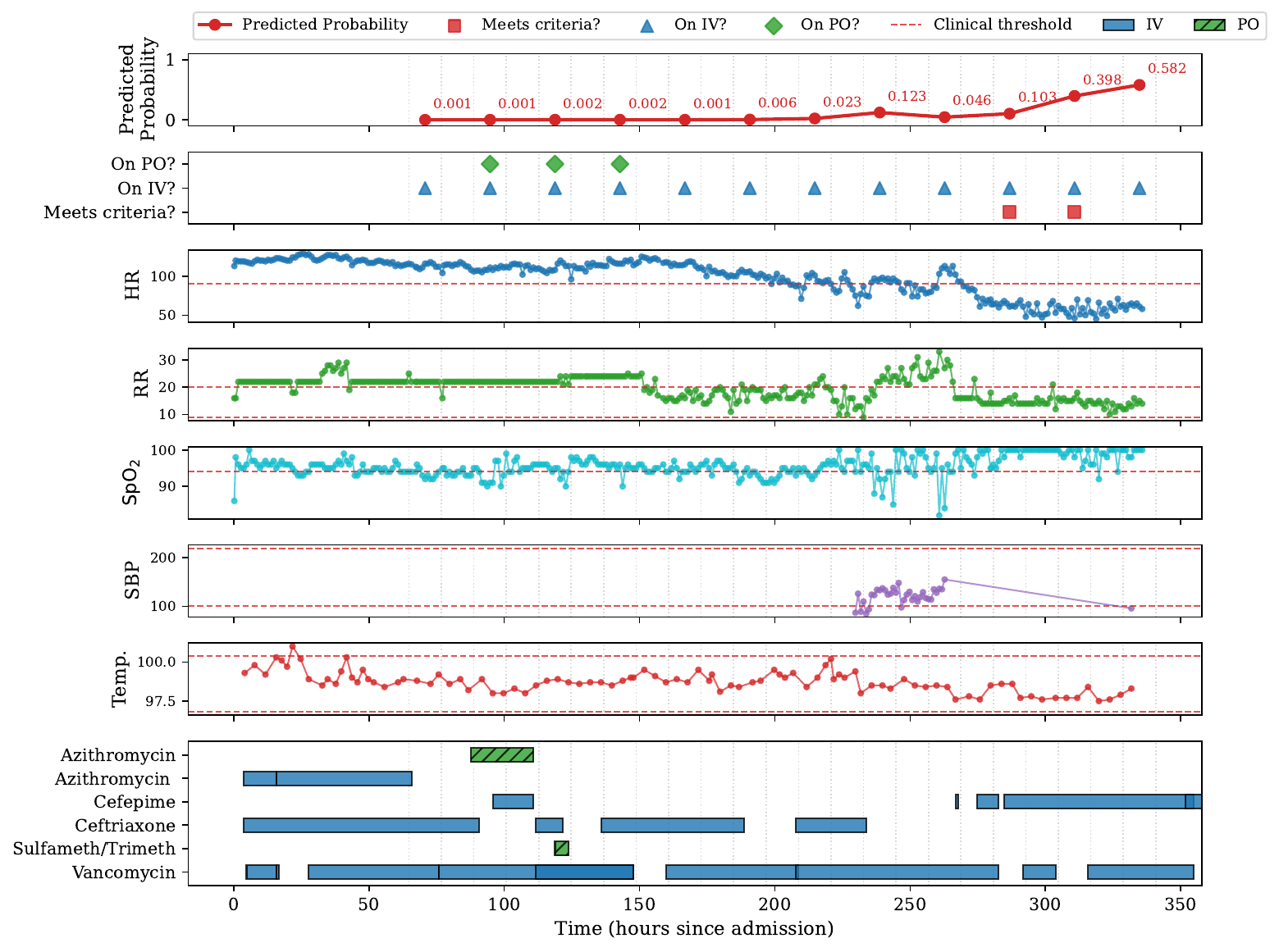}
    \caption{Example encounter from the MIMIC test set illustrating the intended functioning of the system. The panels show (from top to bottom): predicted switch-readiness probabilities from the NP-tuned model; labels indicating oral (PO) and IV antibiotic prescriptions and whether clinical criteria are satisfied in each forecast window; measured vital signs with clinical thresholds for switch-readiness overlaid; and antibiotic prescription timelines, with colour and pattern indicating administration route. Grey vertical lines demarcate daily 12-hour forecast windows. Forecasts begin 48 hours after admission to ensure sufficient historical data for prediction. Variable Abbreviations: HR (Heart Rate), RR (Respiratory Rate), $\text{SpO}_2$ (Oxygen Saturation), SBP (Systolic Blood Pressure), Temp. (Temperature). Antibiotic Abbreviations: Sulfameth/Trimeth (Sulfamethoxazole-Trimethoprim).}
    \label{fig:patient_pathway}
\end{figure}

To illustrate how the proposed system would function in clinical practice, we examine a patient encounter from the test set. \Cref{fig:patient_pathway} shows the progression of this encounter. Two further examples can be found in Supplementary \labelcref{sec:further_example_encounters}. Over the course of approximately two weeks, the patient receives five different antibiotic agents across multiple separate prescriptions. As the encounter progresses, the patient's vital signs steadily improve, particularly heart rate, temperature, and oxygen saturation. The model's predicted switch-readiness probability tracks this clinical trajectory closely: remaining low during the initial period when the patient is more unwell, then rising substantially as physiological stability returns. Notably, despite the patient never being switched to oral therapy in reality, the model identifies the patient as ready before the end of their encounter, suggesting that an earlier switch could have been safely achieved had the proposed system been deployed. 
\par
While this example demonstrates successful model behaviour, it also reveals the inherent complexities of the IVOS problem that the system must address. Vital signs are sampled irregularly, with significant periods of missingness and frequent transient spikes that must be handled robustly. Antibiotic prescribing patterns rarely follow a simple sequence of an IV agent being replaced by its oral equivalent; in practice, multiple drugs may be prescribed concurrently for different indications, making switches difficult to identify from prescription data alone. A prescription of an oral drug does not necessarily indicate the patient's state is improving.
Labelling patients as meeting clinical criteria is similarly challenging: a single measurement marginally exceeding the threshold can cause the label to be negative, as occurs in the final forecast window where the patient's heart rate briefly spikes. Note that it is also possible that two patients may have identical physiological measurements whilst one is suitable for switching and one is not; the underlying indication or issues with tolerance of the oral route might make the switch inappropriate.
\par
Despite these complexities, the model produces clinically sensible predictions that track the patient's physiological trajectory and could support timely identification of switch-ready patients. These challenges motivate the design choices discussed in \Cref{sec:methods}, particularly the aggregation strategy for handling irregular measurements and the choice to forecast vital signs rather than directly predicting switches.

A key contribution of this work is the formulation itself, which differs from previously proposed approaches for the integration of ML for IVOS decision support \citep{bolton_personalising_2024}.  Both the NP and GBDT models produce substantially better results than random selection using our formulation, demonstrating that forecasting physiology can effectively identify patients for IVOS review, independent of the specific model used. The NP and GBDT models perform similarly in terms of ranking performance, which raises the question: why use probabilistic forecasting at all? We stress that our intention is to create a practical machine learning system, as opposed to simply a performant model. Explainability is a critically important factor in the successful adoption of any ML-based CDSS \citep{scottAchievingLargescaleClinician2024}. Clinicians must feel confidence in the suggestions of any model, and have an understanding of how they have been generated. While the NP model itself is not interpretable, the system provides explainability through its outputs and architecture. The system produces forecasts of vital signs into the near future, quantities that clinicians already have a clear understanding of, and applies transparent clinical criteria to these forecasts to determine switch-readiness. When a patient is flagged as switch-ready, clinicians can inspect the underlying vital sign forecasts to understand why.

A further structural advantage of the NP approach is that a single model serves both the forecasting and classification tasks. Because the NP produces a full predictive distribution over future vital sign values, these distributions can be passed directly through the clinical criteria to yield switch-readiness probabilities, which in turn provide a principled ranking of patients for review. The baseline models, by contrast, produce only point estimates when used for forecasting and therefore cannot generate the distributional output needed to assess the probability that a patient's vital signs will remain within normal ranges. As a result, two entirely separate GBDT models are required: one for forecasting and one for classification, each trained on different objectives and feature representations. Beyond the practical burden of maintaining two independent models, this separation means that the classification model's predictions are not grounded in an explicit forecast of the patient's physiological trajectory, weakening the interpretable link between predicted vital signs and the switch-readiness score that clinicians would inspect.
\par
The separation between forecasting and application of criteria allows the model to be used much more flexibly in deployment. Take the example of a patient on mechanical ventilation, which controls respiratory rate, potentially causing a patient to seem ready for switching when they're still unwell. In this case, the measured respiratory rate may fall within normal ranges not because the patient has improved, but because the ventilator is artificially regulating their breathing. When using the forecasting approach, the respiratory rate can be discounted from the switch-readiness calculation very easily, or the threshold can be adjusted on a case by case basis. Likewise if recommended clinical thresholds change with new evidence, the proposed system can be adapted without retraining or adjusting the base model. This is highly beneficial when considering model deployment, since regularly updating and retraining models can add significant complexity \citep{harris_clinical_2022}.

To the best of our knowledge, there is only a single work that discusses the application of ML to the IVOS problem in order to assess the suitability of patients for switching \citep{bolton_personalising_2024}. Their approach is to train a model to predict, for a given day of a patient's encounter, whether that patient received IV antibiotics. By training a model to predict historical clinical actions (i.e., the administration of IV antibiotics), the resulting system will necessarily learn to replicate the existing patterns and frequency of IVOS decisions present in the training data. Clinical practice is changing, and for many conditions there is growing evidence that long courses of IV antibiotics are unnecessary \citep{wald-dickler_oral_2022,li_oral_2019,iversen_partial_2019}. If the goal of a CDSS is to improve upon current practice—in this case by encouraging earlier switches and thus reducing the total number of IV therapy days—then a model trained to mimic historical, suboptimal behaviour is inherently limited. This limitation is particularly acute for the complex or non‑standard cases where clinical decision support is most needed. Although performance of such a model on held-out retrospective data might appear encouraging, this approach inevitably results in a brittle CDSS, frozen in time and incapable of providing continuous support to clinicians participating in a resilient Learning Health System \citep{friedman_what_2022,kilbourne_foundational_2024}. Our approach of modelling patient physiology keeps the prescribing clinician front and centre and results in a CDSS sufficiently flexible to adapt to clinical evidence and changing guidelines.

Fundamentally, obtaining ground truth for switch-readiness is challenging. The decision to switch is multifaceted, involving factors beyond vital signs such as infection type and severity, antibiotic availability and contraindications, patient comorbidities, and social circumstances affecting discharge planning. Historical switching decisions, as recorded in EHRs, reflect suboptimal practice that the system aims to improve rather than replicate, making them unsuitable as ground truth. Even setting aside the issue of learning from prior practice, constructing reliable ground-truth IVOS labels from EHRs is difficult, and label quality caps model performance and clinical utility. \citet{dutey-magni_feasibility_2021} discuss the complexities of reconstructing antibiotic courses from prescription data alone.
\par
The approach to label construction taken by \citet{bolton_personalising_2024} contains methodological issues that result in a dataset that is noisy and systematically biased. IV and oral therapy periods are defined by taking the earliest start time and the latest end time across all antibiotic prescriptions of a given administration route within a single hospital stay, which can cause switches to be labelled incorrectly and systematically excludes more complex, but common, clinical scenarios, biasing the final dataset towards unrealistically simple and linear treatment trajectories. Additionally all antibiotics are treated as interchangeable, with no check to ensure that the oral drug is a clinically appropriate replacement for the IV drug, or that the prescription indications are for the same condition; for example, treatment with broad-spectrum IV antibiotics can cause C. difficile infection \citep{cymbal_management_2024}, and in serious cases the IV course is halted while an oral antibiotic, often vancomycin, is used to control the C. difficile—disregarding indications, this would be labelled as a switch when in fact one has not occurred. 
\par
Our system does not rely on the use of ground-truth switching labels and so is necessarily incomplete, capturing only the physiological improvement component of the decision. The appropriateness of this proxy can only be validated through prospective deployment, where clinicians review flagged patients and provide feedback on the relevance and timing of recommendations. We note that modelling of physiological factors only is by design, our system purposefully keeps the final switch decision in the hands of the clinician. Modelling clinician behaviour and hospital policy is fraught, and leads to models that can reinforce sub-optimal practices. This issue is critically important when considering ML-based CDSSs.

The primary limitation of our work is the absence of prospective clinical evaluation. Without deploying the system in practice and measuring its impact on switching rates, the total number of IV therapy days, patient outcomes, and clinician decision-making, we cannot definitively assess whether the model's predictions translate into improved clinical care. Our evaluation demonstrates that the model can accurately rank patients according to physiological criteria for switch-readiness, but retrospective performance metrics do not guarantee clinical utility. This limitation is shared across the IVOS CDSS literature: to our knowledge, no ML-based system for IVOS has been prospectively evaluated \citep{bolton_personalising_2024}, although \citet{bolton2025impact} conduct a trial which measures the influence of ML-based IVOS recommendations on clinician decision making for a small set of curated case vignettes. 
While rule-based systems have undergone prospective evaluation \citep{beeler_earlier_2015, berrevoets_electronic_2017, kan_implementation_2019}, these studies typically combine the intervention with education and policy changes, making it difficult to isolate the contribution of the CDSS itself.

The model's reliance on five vital signs means that other potentially clinically relevant information is not incorporated. It is possible that better forecasting performance could be achieved by considering a broader range of input variables, for example inflammatory markers (C-reactive protein, white blood cell count), which have been considered in prior work  \citep{10.1609/aaai.v38i14.29534,de2019gru}. The simple set of five vitals were chosen because they are measured with sufficient regularity in the clinical contexts in which the proposed IVOS system could be applied. The forecasting model also assumes that the recent past is informative of the near future, which may not hold during rapid clinical deterioration or improvement, potentially leading to delayed detection of status changes. Finally, as discussed in Supplementary \labelcref{sec:supp_calibration}, the standard NP model exhibits calibration issues that arise from the interaction between the Gaussian likelihood and outliers in EHR data. While the tuned variant addresses this limitation, the need for this correction highlights the difficulty of modelling complex EHR data. 

\section{Methods} \label{sec:methods}
This section details our data sources and preprocessing pipeline, the methodology for constructing switch-readiness labels, the neural process architecture used for forecasting, and the baseline models against which we compare.

\subsection{Data}\label{sec:methods_data}

To train and validate our models, we use two databases of routinely collected EHRs. The first is MIMIC-IV v3.1 \citep{johnson_mimic-iv_2023}, a publicly available, de-identified database of admissions to the intensive care unit (ICU) at the Beth Israel Deaconess Medical Center in Boston, USA, spanning the years 2008-2022. The second is formed from an internal database of admissions to University College London Hospitals (UCLH) NHS Foundation Trust, a group of 11 hospitals in London, UK, spanning the years 2019-2025.

The two datasets represent significantly different populations: comprising only ICU admissions, the MIMIC patients are generally much more unwell, the measurements of patient vital signs are taken much more regularly in the MIMIC data, and the properties of antibiotic prescriptions differ both in terms of route and agents. The datasets are collected from different countries, where care practices, particularly related to antibiotic prescribing vary; characteristics of each dataset are shown in \Cref{tab:cohort_characteristics}. For example, there is a different split of sex (40.7\% female in MIMIC vs 52.6\% in UCLH), and markedly different prescribing patterns: vancomycin is the most common IV antibiotic in MIMIC (65.7\% of admissions) but is rarely used in UCLH, where co-amoxiclav is dominant (46.5\%). Furthermore, the intensity of monitoring differs: while heart rate, respiratory rate, and oxygen saturation are measured approximately twice as frequently in MIMIC compared to UCLH, the difference is less pronounced for systolic blood pressure, and temperature is actually recorded less frequently in MIMIC.

From each database, we extract three categories of information: vital signs, stay information, and antibiotic prescriptions. Vital signs consist of five key physiological measurements: heart rate, respiratory rate, oxygen saturation ($\text{SpO}_2$), systolic blood pressure, and temperature. These are recorded at irregular intervals throughout each patient's stay. Stay information includes admission and discharge times. Antibiotic prescriptions comprise records of antimicrobial orders, including the drug name, route of administration (IV or oral), and the start and end times of each order.

We now formalise the data structure and task definitions. Let $\mathcal{X} = \{X_{1}, X_2, \dots, X_M\}$ be a dataset of $M$ independent hospital encounters. For each encounter $X_{m}$, we have a set of $N_m$ measurements taken from the patient over the course of their stay. Each measurement is one of $C$ variables, in our case the five vital signs described above. We use the triplet representation, so that each encounter can be written as the set $X_{m}=\{ (t_{m, i}, c_{m, i}, x_{m, i})\}_{i=1}^{N_m}$, where $t_{m, i}\in\mathbb{R}^+$ is the time that the measurement of variable $c_{m, i}\in\{1, \dots, C\}$ was taken, and $x_{m, i}\in\mathbb{R}$ is the measured value. Let $X^{\tau}_{m} = \{ (t', c', x') \in X_m | t' < \tau\}$ denote the history of the encounter prior to time $\tau$. The forecasting task is to predict the patient's state at some future time $t>\tau$ given $X_m^\tau$. The classification task is to predict a binary label $y_m^\tau \in \{0, 1\}$ indicating whether the patient meets the clinical criteria for switch-readiness in a forecast window $[\tau, \tau + \Delta]$, given the history $X_m^\tau$. The methodology for determining these labels is discussed in \Cref{sec:methods_labels}.
\par
We apply the following preprocessing steps.
We drop any measurements outside a predefined range of physically plausible values, defined in Supplementary \labelcref{tab:reasonable_values}. We truncate encounters to be a maximum length of 2 weeks. Pediatrics is already excluded from the MIMIC dataset, for the UCLH data, patients aged under 18 are excluded. As a condition of pseudonymisation required to extract the UCLH data, patients aged over 90 were removed. Throughout, we use the relative time since admission to index encounters. Unless otherwise stated, a history of 48 hours is used for model inputs. 
To match the intended deployment of the model, we construct one task per day for each encounter, with the prediction time set to 9 am. A task is created on a given day only if the patient is considered to be on an IV prescription at the prediction time, there are at least 10 measurements in the preceding 48-hour look-back window, and there is at least one measurement in the 12-hour prediction window (9 am to 9 pm). A patient is considered to be on IV medication if they have an active IV antibiotic order at the prediction time, or if an IV order ended within the preceding 36 hours. More sophisticated approaches for reconstructing antibiotic courses do exist \citep{dutey-magni_feasibility_2021}, but were not employed in this study.
\par
To simulate a realistic deployment scenario and prevent look-ahead bias, we split the data temporally to form the test set. For the UCLH data, encounters admitted after 1 January 2024 form the test set. For the MIMIC data, we use a cutoff of 1 January 2019; however, the de-identification process in MIMIC-IV shifts all dates by a random patient-specific offset, which scrambles the chronological order of admissions between patients. We therefore reconstruct approximate admission dates to enable a temporal split, as described in Supplementary \labelcref{sec:data_temporal_split}. Because reconstructed admission dates are approximate the precision@5 metric should be interpreted with more caution for the MIMIC data relative to the UCLH data. A validation set is created by selecting 10\% of non-test data, with the remainder forming the training set. A full breakdown of the number of tasks in each of the training, validation and test sets for each of the datasets is given in \Cref{tab:task_counts}. 
\par
The baseline classification models are trained and evaluated using tasks constructed as described above. However, for training and evaluating the forecasting models, we construct tasks differently to better represent the general forecasting ability of each model rather than the specific switch-readiness prediction setting. For each encounter, we create tasks by uniformly sampling prediction start times $\tau$. For each $\tau$ we use a 48-hour look-back window as context to predict all measurements within a subsequent 12-hour forecast horizon after $\tau$. We sample tasks with a frequency proportional to encounter length, generating one task for every 24 hours of data, so a 7 day encounter would produce 7 tasks with randomly spaced prediction times. This follows standard practice for task construction when training NPs \citep{bruinsma_convolutional_2024}.

\subsection{Labelling switches}\label{sec:methods_labels}

We define switch-readiness labels based on whether a patient's vital signs fall within clinically normal ranges over the forecast window $[\tau, \tau + \Delta]$. For each vital sign $c$, we specify lower and upper thresholds representing the normal range; these thresholds are given in Supplementary \labelcref{tab:clinical_criteria}. A number of different definitions of clinically normal ranges exist in the context of IVOS \citep{harvey_criteria_2023}. We consulted with clinicians at our institution to determine the levels at which they would consider a patient suitable for review, and maintained these levels across both the MIMIC and UCLH datasets. It may be that different levels are appropriate in different contexts; it is simple to adjust these in our framework, the result would be an altered proportion of patients meeting criteria. In order to investigate the sensitivity of our results to a change in the clinical thresholds, we re-ran the MIMIC ranking experiments with less stringent thresholds, resulting in increasing the number of positive (switch-ready) days from 12.2\% to 50.0\%. We found that the top performing model(s) for each of the ranking metrics (AUROC, AP, Brier score, and precision@5) was near identical. For full results see \labelcref{sec:loose_criteria}.

A key consideration in constructing these labels is that the times at which measurements are taken are not known in advance. During model deployment, predictions must be made at fixed evaluation times without knowledge of when future measurements will occur. Evaluating criteria at the actual measurement times would introduce look-ahead bias, since these times are only available retrospectively. To address this, we aggregate measurements within the forecast window onto a fixed grid. We divide the 12-hour forecast window into four 3-hour intervals and compute the median value of each vital sign within each interval. A patient is labelled as switch-ready ($y_m^\tau = 1$) if all aggregated values fall within their respective normal ranges. If no measurements are available for a given vital sign in an interval, that interval is excluded from the criteria check for that variable. This choice to treat missing values as meeting criteria, rather than violating, is a modelling decision that must be clearly communicated to end users; the alternative would result in substantially more imbalanced class distributions, since missing data is common.

\subsection{Neural processes}\label{sec:methods_np}
Neural Processes (NPs) are a type of probabilistic meta-learning model that uses neural networks to parametrise the prediction map from the space of possible datasets to the space of stochastic processes representing the model’s probabilistic predictions at an arbitrary set of target points \citep{garnelo_conditional_2018,bruinsma_convolutional_2024}. While the preceding definition is technically accurate, it is perhaps of limited utility in understanding how NPs can be useful for the forecasting of vital signs. In the parlance of meta-learning, we can view each of the encounters described in \Cref{sec:methods_data} as an individual task, or dataset. The set of encounters is then known as the meta-dataset. We aim to build a model that, when provided with data from an unseen encounter, can make predictions about the values of the patient state.
NPs generally solve this problem by taking the dataset of past tasks and splitting each task into a context set and a target set. The model is then trained to produce well-calibrated probabilistic predictions at a set of target inputs using the information in the context set. In this case the context set is all the measurements of vital signs in the look-back window and the target points are the measurements in the forecast window. 

Convolutional conditional neural processes (ConvCNPs) are a type of NP which encode the property of translation equivariance into the prediction map \citep{vaughan_convolutional_2022}. Translation equivariance implies that if the time values of the data are shifted by a constant time, the model's predictions will also be shifted by that time. Translationally equivariant NPs map to stationary stochastic processes, meaning they are a useful model class for temporal or spatio-temporal problems where stationarity is a reasonable inductive bias \citep{bruinsma_convolutional_2024}. ConvCNPs use a functional encoding based on a kernel smoothing of the input data, followed by processing with a convolutional neural network. A description of the model architecture, training objective, and the derivation of switch-readiness probabilities from the predictive distributions is provided in Supplementary \labelcref{sec:supp_prob_model}.

Our ConvCNP uses a UNet architecture \citep{ronneberger_u-net_2015}, with 4 convolutional layers before and after the bottleneck, and a length scale of 1 hour for the kernel smoothing. At each target time $t$ and for each vital sign $c$ in the forecast window, the model outputs the parameters of a Gaussian predictive distribution; these predictions are conditionally independent given the context data $X_m^\tau$. The model is trained by minimising the negative log-likelihood of the target observations under this predictive distribution. We train for up to 500 epochs with early stopping (patience of 250 epochs), using a batch size of 512. We perform a grid search over learning rate and number of channels, selecting the model with the lowest validation loss. Full hyperparameter settings are given in Supplementary \labelcref{tab:np-hyperparameters}. All models were trained a single NVIDIA A10 or NVIDIA A40 GPU.

To compute switch-readiness probabilities, we consider two approaches. In the standard NP model, for each vital sign we produce predictive distributions at forecast times aligned with the centre of the 3 hour aggregation intervals described in \Cref{sec:methods_labels}. We then compute the probability that the measurement at each point falls within the clinical threshold range using the cumulative distribution function of the Gaussian. The overall switch-readiness probability is then the product of these individual probabilities across all vital signs and time points, under the conditional independence assumption. In the tuned variant (NP-tuned), we add a classification head to the forecasting model. This head aggregates the encoding from the final convolutional layer across the time dimension using a max-pooling operation, then applies a multi-layer perceptron to output a single switch-readiness probability. The classification head is trained with the forecasting model weights frozen. Architecture details for the classification head are given in Supplementary \labelcref{tab:np-tuned-hyperparameters}.

\subsection{Baselines}\label{sec:methods_baslines}

We develop baseline models for both the forecasting and classification tasks. These baselines use tabular models that take as input a fixed-length feature vector derived from the look-back window $X_m^\tau$. For each vital sign $c \in \{1, \dots, C\}$, we extract 16 features from the measurements in $X_m^\tau$: statistical summaries (mean, standard deviation, minimum, maximum, median, count, 25th and 75th percentiles), temporal features (time since the most recent measurement, time span of observations, and measurement frequency), and trend features (mean and standard deviation of successive slopes, total change from first to last observation, and counts of positive and negative changes). If a measurement type is entirely absent for a given task, its corresponding features are imputed with default values. We also compute two global features describing missingness: the count of absent measurement types and an overall completeness ratio. In total there are 82 features for the classification task.

The classification baselines directly predict the switch-readiness label $y_m^\tau$ from the feature vector. We apply both logistic regression, using the scikit-learn implementation \citep{pedregosa_scikit-learn_2011}, and gradient boosted decision trees (GBDT), using the LightGBM implementation \citep{ke_lightgbm_2017}. We perform a grid search over hyperparameters, selecting the model with the highest average precision on a held-out validation set; the hyperparameter search space is given in Supplementary \labelcref{tab:baseline-hyperparameters}. These models serve as a more challenging baseline for the NP-based system, since they directly optimise the classification objective rather than first forecasting vital signs and then applying criteria. 

For the forecasting task, we use two baselines. The repeat baseline simply uses the most recent measured value of each vital sign prior to the forecast window as the prediction for all future time points. The GBDT forecasting baseline is more sophisticated and must account for the irregularity of the target times and the multiple output variables. Each time point in the forecast window is treated as a separate prediction instance. The shared feature vector representing the look-back window is augmented with a one-hot encoding specifying which vital sign $c$ is being predicted, and the time elapsed since the prediction time $\tau$ (i.e., the forecast horizon). This formulation allows a single GBDT model to predict all vital signs at arbitrary future times within the forecast window. Neither baseline produces probabilistic forecasts, limiting their ability to quantify uncertainty in predictions.

\section*{Contribution statement}

M.R. conceived and designed the study. M.R. developed the model architecture, code, and visualisations. S.H., A.L., E.M and N.S. provided clinical guidance and validation. V.L., S.H. and I.J.C. provided modelling expertise and guidance on experiments. S.H. and A.L managed data governance and access approvals. M.R. wrote the initial draft of the manuscript. M.R., V.L., S.H., I.J.C and A.L. provided feedback and edits to the manuscript. V.L, S.H. and I.J.C. provided supervision. All authors approved the final version of the manuscript. 

\section*{Acknowledgments}
M.R., I.J.C., and V.L. are supported by the EPSRC Digital Health Hub for AMR (EP/X031276/1). S.H. and A.L. are
supported by the National Institute for Health and Care Research University College London Hospitals Biomedical Research Centre.

\section*{Code availability}
The computer code used in this research is available at \newline\url{https://github.com/SAFEHR-data/ivos-model-public}.

\section*{Data availability}
The MIMIC-IV dataset can be accessed at \url{https://physionet.org/content/mimiciv/3.1/}.
The UCLH dataset contains personally identifiable data that has not been approved for external release. Researchers wishing to validate or replicate the results on the UCLH dataset would need to be approved for research collaborations with University College London Hospitals NHS Foundation Trust, and to secure appropriate permissions from the UCLH/UCL Joint Research Office. Researchers who meet these requirements can contact the corresponding author for further information about access to the datasets.

\bibliographystyle{unsrtnat} 
\bibliography{refs-clean.bib}

\newpage
\appendix
\input{sup}

\end{document}

%% file: sup.tex
\newcommand{\beginsupplement}{
        \setcounter{section}{0}
        \renewcommand{\thesection}{S\arabic{section}}
        \setcounter{table}{0}
        \renewcommand{\thetable}{S\arabic{table}}
        \setcounter{figure}{0}
        \renewcommand{\thefigure}{S\arabic{figure}}
        \setcounter{equation}{0}
        \renewcommand{\theequation}{S\arabic{equation}}
}

\thispagestyle{empty}
\beginsupplement

\begin{center}
{
\sffamily\LARGE\bfseries Supplementary Information
}
\end{center}

\section{Extended Discussion on Rule-based CDSSs}
\label{sec:supp_rule_based}

There are a number of works that use simple electronic reminders to alert clinicians to review IVOS eligibility after a fixed period of time. \citet{beeler_earlier_2015} implemented a system that alerts clinicians after the patient has been on a course of IV for longer than 60 hours. In order to be eligible to raise an alert the patient must have at least 24 hours left on their prescription, have normal neutrophils and no fever and be able to swallow. They found that this system resulted in a 17\% drop in IV therapy days, in a prospective evaluation. \citet{berrevoets_electronic_2017} built a similar tool that triggers an alert 72 hours after initial prescription if the patient’s inflammation markers are showing signs of improvement, and the patient can tolerate the oral route and does not have an infection type classed as severe. They combine this system with weekly education sessions for clinicians and found a 13\% drop in IV antibiotic usage after 72 hours in a prospective evaluation. \citet{kan_implementation_2019} implement a system that generates an alert for patients with an IV antibiotic order that has been active for at least 48 hours and has no fixed duration, with no fever and an order for an oral diet. They find that this system, paired with a revision of IVOS policy and staff education leads to a significant improvement in switching rate in the post-intervention period.
\citet{quintens_basic_2022} discuss the implementation of a more complex rule based on a set of 13 criteria, with the goal of increasing clinician acceptance of switch alerts. This rule concerns the suitability of oral treatment, and does not directly address the patient's state of clinical improvement. 

The CDSSs discussed above have demonstrated that significant improvements in IVOS rates can be generated by the implementation of simple clinical rules, combined with education of clinicians. The rules discussed in these works primarily address the question of oral treatment suitability, and do not aim to assess the clinical improvement of the patient in a sophisticated way, if at all. In the present work, we aim to use ML to better understand this aspect of the IVOS decision. Note it is not possible to benchmark these rule based methods against the proposed system because we do not have access to reliable ground truth switching labels.

\section{Model Calibration}
\label{sec:supp_calibration}

\begin{figure}[h]
    \centering
    \includegraphics[width=0.7\textwidth]{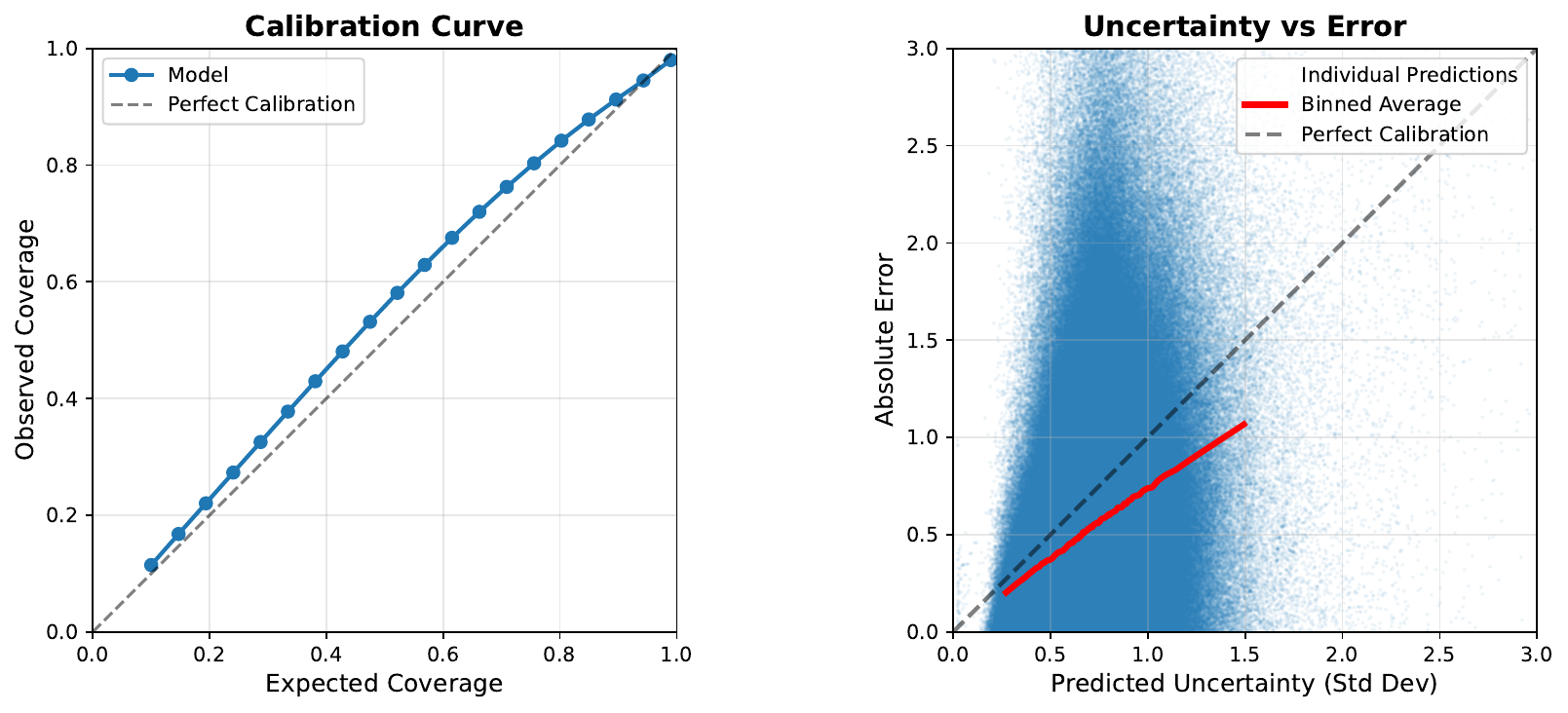}
    \caption{Left plot shows calibration curve for the forecasting model's predicted uncertainty. For each confidence level (e.g., 80\%), we generate a prediction interval that should contain the true value with that probability using the Normal predictive distribution. The x-axis shows the chosen confidence level while the y-axis shows the actual proportion of test observations that fell within the intervals. Perfect calibration follows the diagonal: an 80\% interval should contain 80\% of observations. Points above the diagonal indicate underconfident predictions. Right shows relationship between predicted uncertainty and absolute prediction error. Individual predictions are shown as scatter points, with binned averages over 1st to 99th percentiles (red line) summarising the trend. The diagonal represents ideal calibration where predicted uncertainty magnitude matches actual error magnitude. In this case, an underconfident model has binned average error below the diagonal; where the predicted uncertainty is higher than the actual error.}
    \label{fig:forecasting_calibration}
\end{figure}

As shown in the main text results, the standard NP model exhibits poor calibration for the switch-readiness classification task, despite strong ranking performance. The elevated Brier score indicates that the model's predicted probabilities do not accurately reflect the true likelihood of switch-readiness. Specifically, the model consistently underestimates probabilities of switch-readiness.
\par
This miscalibration stems from the interaction between the training objective and outliers in EHR data. The model minimises the negative log-likelihood of observed measurements under a Gaussian predictive distribution. Because Gaussian distributions have light tails, they assign very low probability to the measurement errors and transient spikes common in routinely collected vital signs. When the model encounters such outliers during training, the NLL becomes very large for predictions with narrow uncertainty, causing the model to learn to widen its predictive distributions. \Cref{fig:forecasting_calibration} demonstrates this through two complementary views: the left panel shows the calibration curve lies above the diagonal, indicating uncertainty estimates are too high, while the right panel shows binned average errors fall below the diagonal, meaning that the realised error of the model's forecasts is consistently lower than the predicted uncertainty.
\par
When an underconfident forecasting model is used to predict switch-readiness by applying clinical criteria to the forecast distributions, this miscalibration propagates to the classification task. The wider predictive distributions increase the probability that forecast values will fall outside the normal ranges that define switch-readiness. This results in predicted switch-readiness probabilities that are systematically too low, producing the poor calibration observed in the results.
\par
The NP-tuned variant addresses this issue by adding a classification head trained directly on the switch-readiness labels. This head learns to correct for the forecasting model's underconfidence, producing well-calibrated probabilities that match the observed frequencies. The tuned model achieves Brier scores comparable to the baseline methods while maintaining strong ranking performance. This approach allows the system to retain the interpretability benefits of the forecasting framework while providing calibrated probability outputs suitable for clinical decision support.

\begin{table}[h!]
\centering
\label{tab:reasonable_values}
\begin{tabular}{lcc}
\toprule
\textbf{Variable} & \textbf{Range} & \textbf{Units} \\
\midrule
Heart Rate & 10 -- 400 & bpm \\
Systolic Blood Pressure & 0 -- 400 & mmHg \\
Oxygen Saturation & 0 -- 100 & \% \\
Respiration Rate & 0 -- 120 & bpm \\
Temperature & 50 -- 120 & °F \\
\bottomrule
\end{tabular}
\caption{Ranges of plausible values used for data cleaning.}

\end{table}

\section{Data setup}

\subsection{Temporal split} 
\label{sec:data_temporal_split}

A key step in our experimental design is a strict temporal split of the data, which is necessary to simulate a realistic deployment scenario and prevent look-ahead bias. The de-identification process in MIMIC-IV shifts all dates for a given patient by a random offset, which preserves intra-patient timelines but scrambles the chronological order of admissions between different patients. A random split on this scrambled data would incorrectly mix past and future data, leading to an over-optimistic evaluation. To address this, we reconstruct an approximate, chronologically-sortable admission date for each encounter. Following the methodology described by the MIMIC-IV creators, we combine the real admission year, available in the \texttt{admissions} table as \texttt{anchor\_year}, with the month and day from the shifted \texttt{admittime}. While this does not recover the exact date, it allows us to correctly order all encounters by year.

\subsection{Number of tasks}

\begin{table}[h]
\centering
\begin{tabular}{lrrrr}
\toprule
 & \multicolumn{2}{c}{\textbf{MIMIC-IV}} & \multicolumn{2}{c}{\textbf{UCLH}} \\
\cmidrule(lr){2-3} \cmidrule(lr){4-5}
\textbf{Partition} & \textbf{Forecasting} & \textbf{Switch-readiness} & \textbf{Forecasting} & \textbf{Switch-readiness} \\
\midrule
Train & 167,923 & 79,148 & 344,133 & 76,586 \\
Val. & 18,804 & 8,902 & 38,864 & 8,885 \\
Test & 52,537 & 22,615 & 186,759 & 41,053 \\
\bottomrule
\end{tabular}
\caption{Number of tasks in each dataset partition for the forecasting and switch-readiness prediction tasks.}
\label{tab:task_counts}
\end{table}

\subsection{Switch criteria}

\begin{table}[h!]
\centering
\label{tab:clinical_criteria}
\begin{tabular}{lc}
\toprule
\textbf{Measurement} & \textbf{Switch Criterion} \\
\midrule
Temperature ($^{\circ}$F) & 96.8 -- 100.4 \\
Heart Rate (bpm) & 41 -- 90 \\
Respiratory Rate (breaths/min) & 9 -- 20 \\
Systolic BP (mmHg) & 101 -- 219 \\
SpO2 (\%) & $> 94$ \\
\bottomrule
\end{tabular}
\caption{Physiological criteria for switch readiness.}

\end{table}

\clearpage
\section{Further neural process details}
\label{sec:supp_prob_model}

In this section, we provide a formal description of the Neural Process (NP) model. We employ a Convolutional Conditional Neural Process (ConvCNP) \citep{vaughan_convolutional_2022, bruinsma_convolutional_2024}, which defines a conditional distribution over target measurements given a context set of observed measurements.

\subsection{Model definition}

Following the notation established in the main text, let a patient encounter be represented by a set of observations $X_m = \{(t_{m,i}, c_{m,i}, x_{m,i})\}_{i=1}^{N_m}$, where $t_{m,i} \in \mathbb{R}^+$ denotes the observation time, $c_{m,i} \in \{1, \ldots, C\}$ indexes the vital sign variable (e.g., heart rate, temperature), and $x_{m,i} \in \mathbb{R}$ is the measured value. Let $X_m^\tau$ denote the history of encounter $m$ prior to time $\tau$, and let $X_m^{[\tau, \tau+\Delta]}$ denote the target observations in the forecast window. The ConvCNP models the predictive distribution $p(X_m^{[\tau, \tau+\Delta]} \mid X_m^\tau)$ through three components:

\textbf{Encoder.} The context set $X_m^\tau$ is mapped onto a functional representation via radial basis function (RBF) kernel smoothing, placing observations onto a discretised temporal grid. This encoding, known as ConvDeepSet \citep{bruinsma_convolutional_2024, jha_neural_2023}, maintains separate channels for data density and signal values for each vital sign variable.

\textbf{Processor.} A UNet convolutional architecture \citep{ronneberger_u-net_2015} processes the gridded representation. Our implementation uses four convolutional blocks before and after the bottleneck layer, with the number of feature channels $D$ as a hyperparameter. Let $\Phi$ denote the composition of encoder and processor. The resulting representation is
\begin{equation}
    \mathbf{R} = \Phi(X_m^\tau) \in \mathbb{R}^{D \times G},
\end{equation}
where $G$ is the number of grid points in the discretised time domain.

\textbf{Decoder.} A convolutional layer with kernel size 1 projects the $D$ feature channels to $2C$ output channels, yielding the mean $\mu_{t,c}$ and variance $\sigma^2_{t,c}$ for each vital sign $c \in \{1, \ldots, C\}$. Parameters at any continuous target time $t$ are recovered by interpolating the projected grid.

\subsection{Training objective}

The Conditional Neural Process framework assumes that target observations are conditionally independent given the context set. The joint predictive likelihood factorises as
\begin{equation}
    p(X_m^{[\tau, \tau+\Delta]} \mid X_m^\tau) = \prod_{(t_j, c_j, x_j) \in X_m^{[\tau, \tau+\Delta]}} \mathcal{N}\bigl(x_j \mid \mu_{t_j, c_j},\, \sigma^2_{t_j, c_j}\bigr),
\end{equation}
where $\mu_{t_j, c_j}$ and $\sigma^2_{t_j, c_j}$ are the decoder outputs for time $t_j$ and variable $c_j$.

The model is trained by minimising the negative log-likelihood (NLL) of the target observations:
\begin{equation}
    \mathcal{L}_{\text{forecast}} = -\frac{1}{|X_m^{[\tau, \tau+\Delta]}|}\sum_{(t_j, c_j, x_j) \in X_m^{[\tau, \tau+\Delta]}} \log \mathcal{N}\bigl(x_j \mid \mu_{t_j, c_j},\, \sigma^2_{t_j, c_j}\bigr).
\end{equation}
This objective encourages accurate point forecasts via the mean while learning uncertainty estimates via the variance that capture the noise inherent in vital signs data.

\subsection{Switch-readiness probability}

Switch-readiness is determined by whether all vital signs remain within clinically defined normal ranges throughout the forecast window $[\tau, \tau + \Delta]$. For each vital sign $c$, let $[\ell_c, u_c]$ denote the acceptable range. Let $\mathcal{T} = \{t_1, \ldots, t_K\}$ be the set of evaluation time points on the forecast grid (at 3-hour intervals), and let $\tilde{X}_{t,c}$ denote the random variable for the predicted measurement of vital sign $c$ at time $t$, distributed as $\tilde{X}_{t,c} \sim \mathcal{N}(\mu_{t,c}, \sigma^2_{t,c})$.

The probability that a patient is switch-ready, denoted $P(y_m^\tau = 1 \mid X_m^\tau)$, is the probability that all forecast measurements satisfy the clinical criteria:
\begin{equation}
    P(y_m^\tau = 1 \mid X_m^\tau) = P\!\left(\bigcap_{t \in \mathcal{T}} \bigcap_{c=1}^{C} \ell_c \leq \tilde{X}_{t,c} \leq u_c \right).
\end{equation}
Under the conditional independence assumption, this joint probability factorises:
\begin{equation}
    P(y_m^\tau = 1 \mid X_m^\tau) = \prod_{t \in \mathcal{T}} \prod_{c=1}^{C} P\bigl(\ell_c \leq \tilde{X}_{t,c} \leq u_c\bigr).
\end{equation}
Each marginal probability is computed analytically using the cumulative distribution function of the Gaussian.

\subsection{Classification head}

For the NP-tuned variant, we augment the model with a classification head to directly predict the binary switch-readiness label $y_m^\tau \in \{0, 1\}$. This head operates on the feature representation $\mathbf{R} \in \mathbb{R}^{D \times G}$ produced by the processor. Global max-pooling over the temporal dimension yields a fixed-size embedding:
\begin{equation}
    h_d = \max_{g \in \{1, \ldots, G\}} R_{d,g}, \quad \mathbf{h} = [h_1, \ldots, h_D]^\top \in \mathbb{R}^D.
\end{equation}
A multi-layer perceptron maps this embedding to a predicted probability $\hat{p} = \text{MLP}(\mathbf{h}) \in [0,1]$.

The classification head is trained in a second stage with the forecasting model weights frozen, using binary cross-entropy loss:
\begin{equation}
    \mathcal{L}_{\text{class}} = -\bigl[y_m^\tau \log \hat{p} + (1 - y_m^\tau) \log (1 - \hat{p})\bigr].
\end{equation}
This two-stage approach calibrates probabilities to the binary task labels while retaining the feature representations learned from the forecasting objective.

\clearpage
\section{Neural Process Hyperparameters}

\begin{table}[h!]
\centering
\begin{tabular}{ll}
\toprule
\textbf{Parameter} & \textbf{Value} \\
\midrule
\multicolumn{2}{l}{\textit{Architecture}} \\
Number of layers & 4 \\
Number of channels$^*$ & \{64, 128, 256, 512\} \\
Lengthscale & 1.0 hour \\
\midrule
\multicolumn{2}{l}{\textit{Training}} \\
Learning rate$^*$ & \{$10^{-5}$, $5 \times 10^{-5}$, $10^{-4}$, $5 \times 10^{-4}$, $10^{-3}$\} \\
Batch size & 512 \\
Epoch size & 16384 tasks \\
Validation epoch size & 4096 tasks \\
Maximum epochs & 500 \\
Early stopping patience & 250 epochs \\
Learning rate scheduler & Cosine with 50 epochs linear warmup \\
\midrule
\multicolumn{2}{l}{\textit{Data}} \\
Lookback window & 48 hours \\
Forecast window & 12 hours \\
Forecast grid size & 3 hours \\
Minimum task measurements & 10 \\
Training tasks per day & 1.0 \\
\bottomrule
\end{tabular}
\caption{Neural process model hyperparameters. Parameters marked with $^*$ were selected via grid search, with the search space shown in parentheses.}
\label{tab:np-hyperparameters}
\end{table}

\begin{table}[h!]
\centering

\begin{tabular}{ll}
\toprule
\textbf{Parameter} & \textbf{Value} \\
\midrule
\multicolumn{2}{l}{\textit{Architecture}} \\
Hidden layer 1 size & 256 \\
Hidden layer 2 size & 128 \\
Dropout rate & \{0.0, 0.1, 0.5\} \\
\midrule
\multicolumn{2}{l}{\textit{Training}} \\
Learning rate & \{$5 \times 10^{-5}$, $10^{-4}$, $5 \times 10^{-4}$, $10^{-3}$\} \\
Other settings & Same as base model \\
\bottomrule
\end{tabular}
\caption{NP-tuned classification head hyperparameters. The classification head is trained with the base forecasting model weights frozen.}
\label{tab:np-tuned-hyperparameters}
\end{table}

\begin{table}[h!]
\centering
\begin{tabular}{ll}
\toprule
\textbf{Parameter} & \textbf{Value} \\
\midrule
\multicolumn{2}{l}{\textit{GBDT}} \\
Number of estimators & \{5, 10, 50, 100, 200\} \\
Maximum depth& \{3, 4, 5, 6\} \\
Learning rate & \{0.001, 0.005, 0.05, 0.1, 0.5\} \\
\midrule
\multicolumn{2}{l}{\textit{Logistic Regression}} \\
Inverse regularization strength (C) & \{0.1, 1.0, 10.0, 100.0, 1000.0\} \\
\bottomrule
\end{tabular}
\caption{Baseline model hyperparameters. A grid search was performed over the sets of hyperparameters. The model with the highest average precision on the validation set was selected for evaluation on the test set.}
\label{tab:baseline-hyperparameters}
\end{table}

\pagebreak

\section{Evaluation Metrics} \label{sec:eval_metrics}

We denote the test set as $\{(\hat{p}_i, y_i)\}_{i=1}^N$, where $\hat{p}_i \in [0,1]$ is the predicted switch-readiness probability and $y_i \in \{0,1\}$ is the true label for instance $i$. For a decision threshold $\theta \in [0,1]$, let $\text{TP}(\theta)$, $\text{FP}(\theta)$, $\text{TN}(\theta)$, and $\text{FN}(\theta)$ denote the number of true positives, false positives, true negatives, and false negatives, respectively.

\paragraph{Area Under the ROC Curve (AUROC)}
The AUROC assesses the model's ability to discriminate between switch-ready and non-switch-ready patients. It represents the probability that a randomly selected positive instance receives a higher predicted probability than a randomly selected negative instance. The AUROC is computed as the integral of the Receiver Operating Characteristic curve, which plots the true positive rate against the false positive rate at varying thresholds:
\begin{equation}
    \text{AUROC} = \int_0^1 \text{TPR}\bigl(\text{FPR}^{-1}(t)\bigr) \, dt,
\end{equation}
where the true positive rate and false positive rate at threshold $\theta$ are defined as
\begin{equation}
    \text{TPR}(\theta) = \frac{\text{TP}(\theta)}{\text{TP}(\theta) + \text{FN}(\theta)}, \qquad
    \text{FPR}(\theta) = \frac{\text{FP}(\theta)}{\text{FP}(\theta) + \text{TN}(\theta)}.
\end{equation}
In practice, this integral is approximated using the trapezoidal rule over the empirical ROC curve.

\paragraph{Average Precision (AP)}
Average Precision summarises the precision-recall curve and is particularly informative for imbalanced classification tasks. Unlike AUROC, which treats positive and negative classes symmetrically, AP focuses on performance with respect to the positive class.

AP is computed as the area under the precision-recall curve, approximated as the weighted sum of precisions at each threshold:
\begin{equation}
    \text{AP} = \sum_{k=1}^{K} \bigl[R(\theta_k) - R(\theta_{k-1})\bigr] \cdot P(\theta_k),
\end{equation}
where $\theta_1 < \theta_2 < \cdots < \theta_K$ are the unique thresholds, and precision and recall are defined as
\begin{equation}
    P(\theta) = \frac{\text{TP}(\theta)}{\text{TP}(\theta) + \text{FP}(\theta)}, \qquad
    R(\theta) = \frac{\text{TP}(\theta)}{\text{TP}(\theta) + \text{FN}(\theta)}.
\end{equation}
A high AP indicates that the model achieves both high recall and high precision.

\paragraph{Brier Score}
The Brier score evaluates the accuracy and calibration of probabilistic predictions. It is defined as the mean squared error between predicted probabilities and observed outcomes:
\begin{equation}
    \text{Brier} = \frac{1}{N} \sum_{i=1}^{N} (\hat{p}_i - y_i)^2.
\end{equation}
A lower Brier score indicates better calibrated and more accurate probability estimates. The score ranges from 0 (perfect predictions) to 1 (maximally incorrect). A model that always predicts the class prior $p$ achieves a Brier score of $p(1-p)$.

\paragraph{Precision@5}
Precision@5 measures clinical utility by evaluating the model's ability to prioritise patients for review. For each day $d$ in the test set with at least 10 active encounters, patients are ranked by predicted switch-readiness probability in descending order. Let $\mathcal{D}$ denote the set of valid days, and let $\pi_d(j)$ denote the index of the patient with the $j$-th highest predicted probability on day $d$. Precision@5 is the proportion of truly switch-ready patients among the top 5, averaged over all valid days:
\begin{equation}
    \text{Precision@5} = \frac{1}{|\mathcal{D}|} \sum_{d \in \mathcal{D}} \frac{1}{5} \sum_{j=1}^{5} y_{\pi_d(j)}.
\end{equation}

\section{Additional results}

\subsection{Criteria threshold sensitivity}

\label{sec:loose_criteria}

\begin{table}[h!]
\centering
\label{tab:clinical_criteria_loose}
\begin{tabular}{lc}
\toprule
\textbf{Measurement} & \textbf{Switch Criterion} \\
\midrule
Temperature ($^{\circ}$F) & 96.8 -- 100.58 \\
Heart Rate (bpm) & 40 -- 131 \\
Respiratory Rate (breaths/min) & 8 -- 24 \\
Systolic BP (mmHg) & 90 -- 229 \\
SpO2 (\%) & $> 91$ \\
\bottomrule
\end{tabular}
\caption{Alternative, loose criteria for switch readiness.}
\end{table}

We tested the sensitivity of the proposed system to a change in clinical criteria since it is likely that different institutions will have different requirements. \Cref{tab:clinical_criteria_loose} shows an alternative looser set of criteria. Applying these criteria to the MIMIC data resulted in an increase in the number of positive (switch-ready) days from 12.2\% to 50.0\%. The relative performance of our model was preserved when measured with the looser criteria. Results for the ranking metrics are shown in \cref{tab:mimic_loose_ranking} and for the binary metrics in \cref{tab:mimic_loose_binary}.

\begin{table}[ht]
\centering
\resizebox{\columnwidth}{!}{
\begin{tabular}{llrrrr}
\toprule
\textbf{Dataset} & \textbf{Model} & \multicolumn{4}{c}{\textbf{Metric}} \\
\cmidrule(lr){3-6}
& & \textbf{AUROC} & \textbf{AP} & \textbf{Brier Score} & \textbf{Precision@5} \\
\midrule
\multirow{4}{*}{\textbf{MIMIC-loose}} & \textbf{Logistic} & 0.676 [0.669-0.682] & 0.650 [0.640-0.660] & 0.229 [0.228-0.231] & 0.680 [0.663-0.696] \\
& \textbf{GBDT} & 0.829 [0.824-0.834] & \bftab 0.809 [0.801-0.817] & \bftab 0.168 [0.165-0.170] &\bftab 0.835 [0.821-0.849] \\
& \textbf{NP} & \bftab 0.836 [0.831-0.841] & \bftab 0.810 [0.802-0.817] & 0.239 [0.235-0.243] & \bftab 0.841 [0.827-0.855] \\
& \textbf{NP-tuned} & 0.814 [0.809-0.820] & 0.792 [0.783-0.800] & 0.175 [0.172-0.177] & 0.822 [0.808-0.838] \\
\bottomrule
\end{tabular}
} 
\caption{Model performance for switch-readiness prediction on the MIMIC dataset with looser clinical criteria (switch rate 50.0\%). Bold values indicate the best performing model(s) (p $\ge$ 0.05). Model abbreviations: GBDT (Gradient Boosted Decision Trees), NP (Neural Process).}
\label{tab:mimic_loose_ranking}
\end{table}
 
\begin{table}[ht]
\centering
\resizebox{\columnwidth}{!}{
\begin{tabular}{llrrrr}
\toprule
\textbf{Dataset} & \textbf{Model} & \multicolumn{4}{c}{\textbf{Metric}} \\
\cmidrule(lr){3-6}
& & \textbf{F1 Score} & \textbf{Accuracy} & \textbf{Precision} & \textbf{Recall} \\
\midrule
\multirow{5}{*}{\textbf{MIMIC-loose}} & \textbf{Repeat} & 0.739 [0.733-0.745] & 0.700 [0.695-0.706] & 0.654 [0.647-0.662] & 0.849 [0.842-0.856] \\
& \textbf{Logistic} & 0.684 [0.678-0.690] & 0.560 [0.554-0.567] & 0.534 [0.527-0.541] & \bftab 0.950 [0.946-0.954] \\
& \textbf{GBDT} & 0.772 [0.766-0.777] & 0.741 [0.736-0.747] & 0.690 [0.683-0.697] & 0.875 [0.869-0.881] \\
& \textbf{NP} & \bftab 0.779 [0.773-0.784] & \bftab 0.753 [0.748-0.759] & \bftab 0.707 [0.699-0.714] & 0.867 [0.861-0.872] \\
& \textbf{NP-tuned} & 0.762 [0.756-0.768] & 0.730 [0.724-0.736] & 0.681 [0.674-0.689] & 0.865 [0.859-0.872] \\
\bottomrule
\end{tabular}
} 
\caption{Model performance for binary switch-readiness prediction on the MIMIC dataset with looser clinical criteria (switch rate 50.0\%). Bold values indicate the best performing model(s) (p $\ge$ 0.05). Model abbreviations: GBDT (Gradient Boosted Decision Trees), NP (Neural Process).}
\label{tab:mimic_loose_binary}
\end{table}

\clearpage
\subsection{Example encounters}
\label{sec:further_example_encounters}
\begin{figure}[h]
    \centering
    \includegraphics[width=1.0\textwidth]{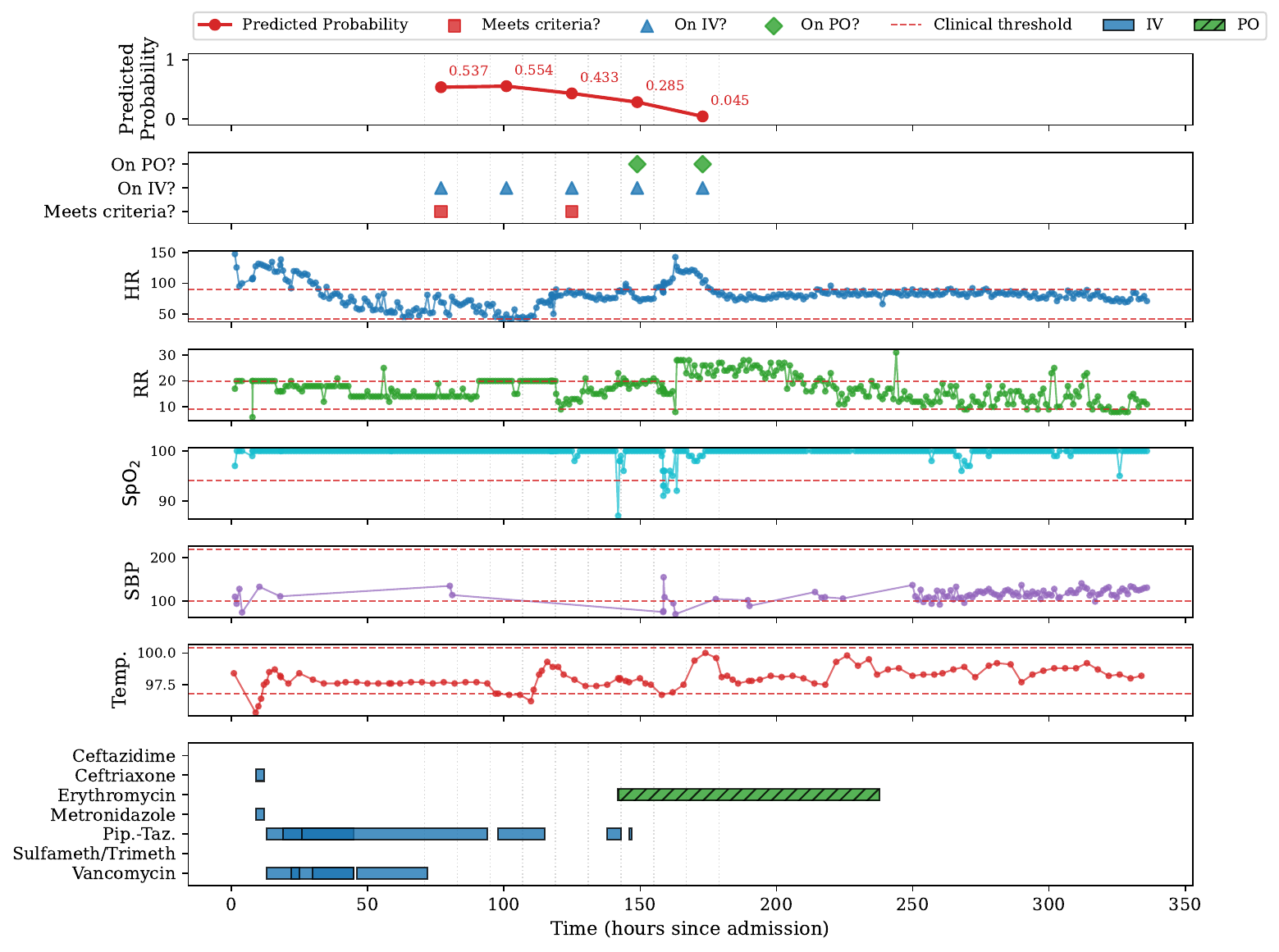}
    \caption{Example encounter from the MIMIC test set illustrating a case in which the system's prediction is less clearly beneficial. The system recommends an earlier switch, but then the patient becomes more unwell as they are switched. In this encounter the test days only span around half of the full stay, since the IV prescription is halted after 150 hours. Panels show the same information as those in the figure in the main text. Variable Abbreviations: HR (Heart Rate), RR (Respiratory Rate), $\text{SpO}_2$ (Oxygen Saturation), SBP (Systolic Blood Pressure), Temp. (Temperature). Antibiotic Abbreviations: Pip.-Taz., Piperacillin-Tazobactam; Sulfameth/Trimeth, Sulfamethoxazole-Trimethoprim}
    \label{fig:patient_pathway2}
\end{figure}

\begin{figure}
    \centering
    \includegraphics[width=1.0\textwidth]{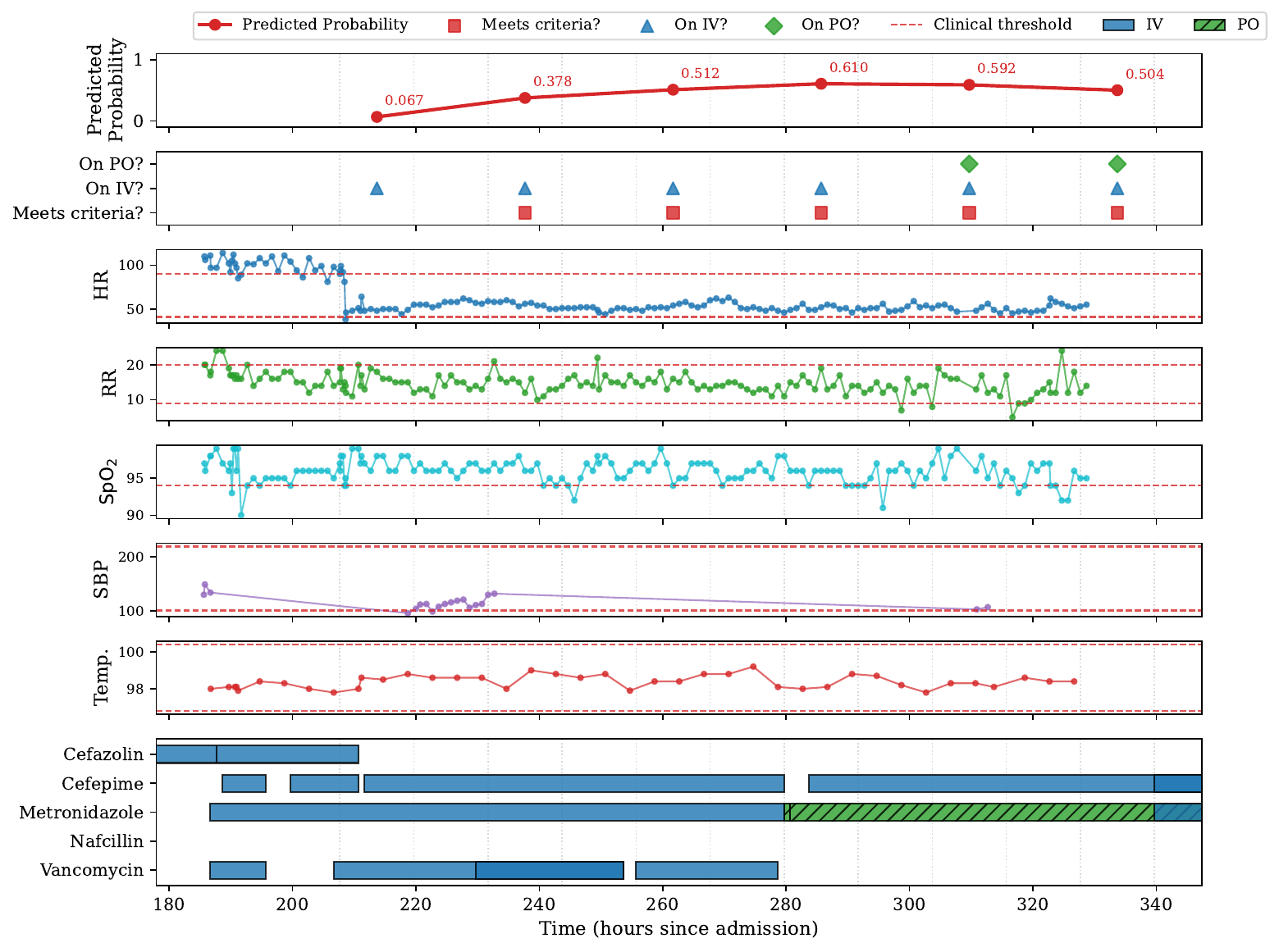}
    \caption{Example encounter from the MIMIC test set. In this example the patient is unwell at the very start of the encounter, but their vitals are stable for most of the remainder, aside from a few spikes. The model becomes more confident the patient is ready to be switched as the encounter progresses, although no switch occurs.  Panels show the same information as those in the figure in the main text.  Variable Abbreviations: HR (Heart Rate), RR (Respiratory Rate), $\text{SpO}_2$ (Oxygen Saturation), SBP (Systolic Blood Pressure), Temp. (Temperature).}
    \label{fig:patient_pathway3}
\end{figure}
\pagebreak
\section{Statistical Significance Testing}
\label{sec:supp_significance}

We compared model performance using paired bootstrap resampling with bootstrap-based hypothesis testing and a Bonferroni correction. For each of 1,000 bootstrap iterations, a single set of indices was generated by sampling with replacement from the test set. These same indices were used to select the predictions for every model, ensuring that performance metrics were calculated on identical subsets of the data for each iteration.

Pairwise comparisons against the best-performing model were conducted using bootstrap difference tests. For each comparison, we calculated the proportion of bootstrap iterations in which the best model outperformed the comparison model. A model was considered significantly different if this proportion exceeded $1 - \alpha/2$ (indicating the best model consistently outperformed the comparison model across bootstrap samples). We applied a Bonferroni correction for multiple testing: for $k$ models, the significance threshold was adjusted to $\alpha/(k-1)$, with $\alpha=0.05$. Models not significantly different from the top performer under this threshold were designated as ``significantly best'' and are indicated with bold in the results tables.